\documentclass{article}
\usepackage[utf8]{inputenc}

\title{On Active Learning for Gaussian Process-based Global Sensitivity Analysis}
\author{Mohit Chauhan, Mariel Ojeda-Tuz, Ryan Catarelli, Kurtis Gurley,\\ Dimitrios Tsapetis, Michael D. Shields}
\date{\vspace{-2ex}}
\usepackage[numbers]{natbib}
\usepackage{graphicx}
\usepackage{dsfont}
\usepackage{multirow}
\usepackage{mathtools}
\usepackage{amsfonts}
\usepackage{xcolor}
\usepackage{wasysym}
\usepackage{hyperref}
\usepackage{bm}
\usepackage[margin=1in]{geometry}
\usepackage[noabbrev]{cleveref}
\DeclareMathOperator*{\argmax}{argmax} 

\begin{document}

\maketitle

\section*{Abstract}
This paper explores the application of active learning strategies to adaptively learn Sobol indices for global sensitivity analysis. We demonstrate that active learning for Sobol indices poses unique challenges due to the definition of the Sobol index as a ratio of variances estimated from Gaussian process surrogates. Consequently, learning strategies must either focus on convergence in the numerator or the denominator of this ratio. However, rapid convergence in either one does not guarantee convergence in the Sobol index. We propose a novel strategy for active learning that focuses on resolving the main effects of the Gaussian process (associated with the numerator of the Sobol index) and compare this with existing strategies based on convergence in the total variance (the denominator of the Sobol index). The new strategy, implemented through a new learning function termed the MUSIC (minimize uncertainty in Sobol index convergence), generally converges in Sobol index error more rapidly than the existing strategies based on the Expected Improvement for Global Fit (EIGF) and the Variance Improvement for Global Fit (VIGF). Both strategies are compared with simple sequential random sampling and the MUSIC learning function generally converges most rapidly for low-dimensional problems. However, for high-dimensional problems, the performance is comparable to random sampling. The new learning strategy is demonstrated for a practical case of adaptive experimental design for large-scale Boundary Layer Wind Tunnel experiments. 

Keywords:  Sobol Index, Active Learning, Global Sensitivity Analysis, Gaussian Process Regression, Kriging

\section{Introduction}
Active learning is widely used in computational physics-based modeling, and increasingly in experimental studies \cite{shields2023}, to serve a variety of objectives. In these schemes, a machine learning (ML) model is employed in an iterative routine to extract information from previous simulations/experiments and identify, through some \textit{learning function} the most informative simulation/experiment to run next. This has been especially beneficial in the fields of optimization and reliability, although it has also found wider applications in uncertainty quantification (UQ) more generally. 

In optimization, active learning is at the heart of Bayesian Optimization (BO). In BO, Bayes Rule is used to define a learning function (termed an acquisition function in BO) that seeks to identify the minimum (or maximum) of some function. Comprehensive reviews of BO can be found in the following references \cite{shahriari2015taking, frazier2018bayesian, greenhill2020bayesian}. A detailed discussion of BO is beyond the scope of this work, but the most common approach in these methods is to approximate the function with a Gaussian Process (GP) and construct a learning function that exploits the jointly Gaussian nature of the function to adaptively learn where the sought extremum has the highest probability of lying. Among these developments, the most widely used learning functions are the Expected Improvement Function (EIF) \cite{movckus1975bayesian, jones1998efficient, huang2006global} the Knowlege Gradient \cite{frazier2009knowledge, scott2011correlated, wu2016parallel} and the Entropy Search \cite{hennig2012entropy, hernandez2014predictive} algorithms.

For reliability analysis, the objective is to estimate the probability of failure of a physical system, which involves solving a high-dimensional probability integral (expectation) over a binary classifier that indicates either safety or failure. This integral is often solved statistically using Monte Carlo methods wherein the integral is estimated as the statistical expectation of the failure indicator by sampling from the input distribution and assessing the indicator at each sample point. Active learning is then used to approximate this indicator function efficiently to minimize the computational burden associated with assessing system failure at each sample with an expensive computational model. Numerous learning functions have been developed for this task, with each attempting to efficiently approximate the limit surface (surface that separates safe and failure domains) in different ways, but in most cases using a GP. The first such method, termed the Expected Feasibility Function (EFF) was proposed by Bichon et. al in 2008 \cite{bichon2008efficient}. This was followed by the widely used U-function developed by Echard et al. \cite{echard2011ak} in the Adaptive Kriging with Monte Carlo Simulation (AK-MCS) method. Since then, various modifications have been made and these concepts have been integrated with other reliability methods (such as subset simulation \cite{huang2016assessing, zhang2019active, xu2020ak}) to improve the efficiency, convergence, and accuracy of these methods \cite{sundar2019reliability, lelievre2018ak, razaaly2020extension, el2021improved}.

Although optimization and reliability are the most common application domains for active learning, it has been applied more broadly in UQ, for example to efficiently construct surrogate models (aka metamodels or emulators) for expensive physics-based computational models. Lam \cite{lam2008sequential}, for example, proposed the Expected Improvement for Global Fit (EIGF) learning function to improve the convergence of GP surrogate models for UQ purposes. Active learning has also been applied using other ML surrogate model forms, such as Polynomial Chaos Expansions (PCE) \cite{marelli2018active, cheng2020active, novak2023active} and Deep Neural Networks (DNNs) \cite{sener2017active,haut2018active, schroder2020survey}. Furthermore, recent efforts have integrated active learning concepts into the design of physical experiments for wind tunnel experiments\cite{deloach1998applications, hill2011examining, vandercreek2019experimental, shields2023}, materials discovery \cite{lookman2019active, kusne2020fly, jablonka2021bias}, and biological sciences \cite{REKER201973, warmuth2003active, naik2016active, sverchkov2017review}.

Regardless of their application or objective, a common challenge in active learning is to balance \textit{exploration} and \textit{exploitation}. That is, any active learning method (i.e.\ learning function) must simultaneously use (exploit) the information obtained from prior simulations/experiments to guide further analyses and, at the same time, explore new regions of the parameter space where little or no data has been collected and uncertainty is correspondingly high. Balancing these objectives is a primary challenge in active learning \cite{osugi2005balancing}, especially in high-dimensional parameter spaces where exploration can be particularly difficult \cite{shan2010survey}. 

In this work, we explore the potential for active learning for Global Sensitivity Analysis (GSA), which to the authors' knowledge, has not previously been studied. Active learning for GSA is of particular interest here because of its potential to aid in other UQ studies. In particular, GSA can be used as an initial filter in UQ studies to reduce the number of random variables that need to be considered in the study. For example, in Section \ref{sec:blwt} we consider a practical boundary layer wind tunnel (BLWT) experimental design problem that has 10 input random variables and, through active learning, determine that only 3 random variables contribute appreciably to the variance of the response. We can then practically ignore uncertainty in the other 7 random variables, thus making UQ more computationally tractable in this complex physical setting. Hence, active learning for GSA allows us to adaptively reduce the dimension rapidly \textit{on-the-fly} and therefore speed up the overall UQ task. 

We show that active learning for GSA presents some unique challenges that do not arise in other active learning domains. These challenges arise from the definition of Global Sensitivity Indices (specifically Sobol indices \cite{sobol1993sensitivity}) as the ratio of two variances, which makes the construction of a learning function that directly targets the Sobol index difficult to derive. Instead, we explore two different learning functions -- one that targets efficient learning for the numerator and one that targets the denominator. We show that each can be beneficial under certain circumstances, but neither approach universally outperforms random sampling. We explore the reasons for this, and specifically demonstrate that this results from the challenges of minimizing the error in a ratio. We discuss when the different approaches appear to be beneficial using a series of analytical functions of varying dimensions. We then demonstrate our use of active learning for GSA on the BLWT application of interest. 

\section{Preliminaries}

Here, we briefly review the foundations of Global Sensitivity Analysis using Sobol Indices and Gaussian Process (GP) regression, upon which all further analyses will depend.

\subsection{Sobol Sensitivity Indices}

Variance-based methods for GSA aim to decompose the variance of the model output $y(\bm{X})$ into distinct contributions from each of the individual random variables $X_i$ (main effects) and the interaction of multiple random variables. In the Sobol method \cite{sobol1993sensitivity, saltelli1995use, archer1997sensitivity}, this is done by first constructing the high-dimensional model representation (HDMR) of the model as:
\begin{equation}
    y(\bm{X}) = f_0 + \sum_{i=1}^d f_i(X_i) + \sum_{i=1}^d \sum_{j>i}f_{ij}(X_i, X_j) + \dots
    \label{eqn:HDMR}
\end{equation}
where 
\begin{equation}
\begin{aligned}
    f_0 &= E[Y]\\
    f_i &= E[Y|X_i] - E[Y]\\
    f_{ij} &= E[Y|X_i,X_j] - f_i - f_j - E[Y]
\end{aligned}
\end{equation}
Taking the variance of Eq.\ \eqref{eqn:HDMR} and exploiting the orthogonality between terms in the expansion yields the following variance decomposition:
\begin{equation}
\begin{aligned}
    \text{Var}(Y) &= \sum_{i=1}^d\text{Var}_{X_i}(E_{X_{\sim i}}[Y|X_i]) + \sum_{i=1}^d \sum_{j>i} \text{Var}_{X_{ij}}(E_{X_{\sim ij}}[Y|X_i, X_j]) + \dots\\
    V(Y) & = \sum_{i=1}^d V_i + \sum_{i=1}^d \sum_{j>i} V_{ij} + \dots
\end{aligned}
\label{eqn:Var_decomp}
\end{equation}
where $V_i$ represent the main effect variances and $V_{ij}$ represent the variance contribution from interactions between variables $X_i$ and $X_j$. The Sobol sensitivity indices reflect the relative contribution of each term in Eq.\ \eqref{eqn:Var_decomp} to the total variance and are given by:
\begin{equation}
    S_i = \dfrac{V_i}{V(Y)} = \dfrac{\text{Var}_{X_i}(E_{X_{\sim i}}[Y|X_i])}{\text{Var}(Y)}
    \label{eqn:main_effect_SI}
\end{equation}
\begin{equation}
    S_{ij} = \dfrac{V_{ij}}{V(Y)} = \dfrac{\text{Var}_{X_{ij}}(E_{X_{\sim ij}}[Y|X_i, X_j]) - \text{Var}_{X_i}(E_{X_{\sim i}}[Y|X_i]) - \text{Var}_{X_j}(E_{X_{\sim j}}[Y|X_j])}{\text{Var}(Y)}
\end{equation}
where $S_i$ are the main effect sensitivity indices and $S_{ij}$ are the interaction sensitivities and we have that 
\begin{equation}
    \sum_{i=1}^d S_i + \sum_{i=1}^d \sum_{j>i} S_{ij} + \dots = 1
\end{equation}
Here, we will focus predominantly on the main effect sensitivity indices in Eq.\ \eqref{eqn:main_effect_SI}, with some supporting discussion of interaction sensitivities. Importantly, we highlight that the sensitivity index $S_i$ is defined as a ratio of variances (the so-called main effect variance in the numerator and total variance in the denominator), which \textit{both} must be estimated in an active learning framework.

\subsection{Gaussian Process Regression / Kriging}
Gaussian Process (GP) regression is a machine learning (ML) method that seeks to identify the best fit Gaussian stochastic process to a set of data points \cite{rasmussen2006gaussian}. Kriging \cite{matheron1963principles, cressie2015statistics, martin2005use}, a variation of GP regression, is a special case wherein the GP regressor serves to interpolate between the set of training points. 
Kriging is a widely-used ML method to construct approximate surrogate models for complex computational models. It is particularly attractive for active learning because it provides a model predictor as well as a measure of the uncertainty in the prediction of the Gaussian posterior standard deviation. 

Consider a computational model $y(\bm{x})$ having input vector $\bm{x}=\{x_1, x_2, \dots, x_d\}\in\mathbb{R}^d$. Next, consider that we have $n$ realizations of $\bm{x}$, denoted by $\bm{X}=\{\bm{x}^{(1)},\bm{x}^{(2)},\dots,\bm{x}^{(n)} \}$ and corresponding model outputs $\bm{Y}=\{y^{(1)},y^{(2)},\dots,y^{(n)} \}$ evaluated by $y^{(i)}=y(\bm{x}^{(i)}$). The pairs, $\bm{X},\bm{Y}$ serve as the training data for the Gaussian process regression wherein the model $y(\bm{x})$ is approximated by 
\begin{align}
\label{eq:}
    \mathcal{Y}(\bm{x}, \omega) = \mathcal{F}(\bm{x}) + Z(\bm{x}, \omega)
\end{align}
where $\mathcal{F}(\cdot)$ is a regression model and $Z(\cdot)$ is a zero-mean Gaussian random process having sample space indexed by $\omega\in\Omega$. We consider that the regression model $\mathcal{F}(\cdot)$ is defined through a linear combination of basis functions $\bm{f}(\bm{x})=\{f_1(\bm{x}),f_2(\bm{x}),\dots, f_p(\bm{x}) \}$ having coefficients $\boldsymbol{\beta}=\{\beta_1, \beta_2, \dots, \beta_p\}$ as
\begin{equation}
    \mathcal{F}(x) = \boldsymbol{\beta}^T \bm{f}(\bm{x})
\end{equation}
The Gaussian random process $Z(\cdot)$ is considered to have zero mean and covariance:
\begin{align}
\label{eq:}
    \mathbb{E}[Z(\bm{x}_1)Z(\bm{x}_2)] = \sigma^2_{z} \mathcal{R}(\bm{x}_1, \bm{x}_2|\boldsymbol{\theta})
\end{align}
where $\mathcal{R}(\cdot)$ is the autocorrelation function having hyperparameters $\boldsymbol{\theta}$. In this work, we will consider the Gaussian correlation model having form 
\begin{equation}
    \mathcal{R}(\bm{x}_1,\bm{x}_2|\boldsymbol{\theta}) = \prod_{k=1}^d \exp(-\theta_k|x_{2k}-x_{1k}|^2 ) = \prod_{k=1}^d r_k(x_{1k},x_{2k}|\theta_k).
    \label{eqn:correlation}
\end{equation}

The GP is then fit to the normalized training data by identifying the posterior distribution of the random process given the data. Given the Gaussian assumption, the posterior distribution is determined by considering that the joint distribution of the GP predictions $\mathcal{Y}(\bm{x})$ and the observations $\bm{Y}$ is a multivariate Gaussian and can be expressed as 
\begin{equation}
    \left\{\begin{array}{c}
         \mathcal{Y}(\bm{x})  \\
         \bm{Y} 
    \end{array}\right\} \sim
    N \left( \left\{
    \begin{array}{c}
         \bm{f}(\bm{x})^T  \boldsymbol{\beta}  \\
         \bm{F}  \boldsymbol{\beta} 
    \end{array}
    \right\}, \sigma_z^2 \left\{
    \begin{array}{cc}
         1 & \bm{r}^T(\bm{x})  \\
         \bm{r}(\bm{x}) & \bm{R}
    \end{array}
    \right\}
    \right)
\end{equation}
where $\bm{F}$ is the matrix of basis function evaluations at the training points given by $F_{ij}=f_j(\bm{x}^{(i)}), i=1,\dots, n, j =1\dots p$, $\bm{r}(\bm{x})$ is the vector of correlations between the prediction point $\bm{x}$ and the training points $\bm{x}^{(i)}$ given by $\bm{r}_i=\mathcal{R}(\bm{x},\bm{x}^{(i)}|\boldsymbol{\theta}), i=1,\dots,n$, and $\bm{R}$ is the correlation matrix of points in the training set given by $R_{ij}=\mathcal{R}(\bm{x}^{(i)},\bm{x}^{(j)}|\boldsymbol{\theta}), i,j=1,\dots,n$.

This posterior can be estimated by identifying the hyperparameters $\boldsymbol{\theta}$, variance $\sigma_z$, and the regression coefficients $\boldsymbol{\beta}$. This is typically done by solving for the parameters that either maximize the likelihood of the observations or minimize the cross-validation error. We apply maximum likelihood estimation where the likelihood function follows from the Gaussian assumption as
\begin{equation}
    \mathcal{L}(\boldsymbol{\theta},\sigma_z,\boldsymbol{\beta}|\bm{Y}) = \dfrac{(\det\bm{R})^{-1/2}}{(2\pi\sigma_z^2)^{n/2}} \exp\left[-\dfrac{1}{2\sigma_z^2}(\bm{Y}-\bm{F}\boldsymbol{\beta})^T\bm{R}^{-1}(\bm{Y}-\bm{F}\boldsymbol{\beta}) \right]
\end{equation}

After estimating the hyperparameters, the mean and the variance of the Gaussian random variable conditioned on the training data are given by:
\begin{equation}
    \hat{y}(\bm{x}) = \bm{f}(\bm{x})^T \boldsymbol{\beta} + \bm{r}(\bm{x})^T \bm{R}^{-1} (\bm{Y} - \bm{F}\boldsymbol{\beta})
    \label{eqn:predictor}
\end{equation}
\begin{equation}
    \sigma^2_{\hat{y}}(\bm{x}) = \sigma^2_z \left(1-\bm{r}(\bm{x})^T \bm{R}^{-1} \bm{r}(\bm{x}) + \bm{t}(\bm{x})^T( \bm{F}^T \bm{R}^{-1} \bm{F})^{-1} \bm{t}(\bm{x}) \right)
    \label{eqn:variance}
\end{equation}
where
\begin{equation}
    \bm{t}(\bm{x}) = \bm{F}^T \bm{R}^{-1}\bm{r}(\bm{x}) - \bm{f}(\bm{x})
    \label{eqn:krig_variable_t}
\end{equation}
These provide the Kriging prediction at a new point $\hat{y}(\bm{x})$ along with a measure of its uncertainty $\sigma_{\hat{y}}(\bm{x})$. 

\section{Estimating Sobol Indices from Gaussian Process Regression}
In this section, we review the process through which Sobol sensitivity indices can be estimated from a GP regression model as originally derived in \cite{MARREL2009742}.
Marrel et al.\ \citep{MARREL2009742} presented two approaches to compute Sobol indices using GPs. In the first approach, the Sobol indices are computed simply using the GP predictor, in which case the method provides a deterministic point estimator based purely on the conditional mean as 
\begin{equation}
    S_i = \dfrac{\text{Var}_{X_i}(E_{X_{\sim i}}[E[\mathcal{Y}(X,\omega)]|X_i])}{\text{Var}(E[\mathcal{Y}(X,\omega)])} = \dfrac{\text{Var}_{X_i}(E_{X_{\sim i}}[\hat{y}(X)|X_i])}{\text{Var}(\hat{y}(X))}
\end{equation}
where $\hat{y}(X)$ denotes the conditional mean in Eq.\ \eqref{eqn:predictor} and is treated as a deterministic function in this approach.

The second approach, which we will employ in this study, uses the full GP model as a stochastic function, which results in random variable sensitivity indices given by:
\begin{equation}
    \tilde{S}_i(\omega) = \dfrac{\text{Var}_{X_i}(E_{X_{\sim i}}[\mathcal{Y}(X,\omega)|X_i])}{E[\text{Var}(\mathcal{Y}(X,\omega))]}
    \label{eqn:Sobol_RV}
\end{equation}
having mean and variance given by:
\begin{equation}
    \mu_{\tilde{S}_i} = \dfrac{E[\text{Var}_{X_i}(E_{X_{\sim i}}[\mathcal{Y}(X,\omega)|X_i])]}{E[\text{Var}(\mathcal{Y}(X,\omega))]}
\end{equation}
and 
\begin{equation}
    \sigma^2_{\tilde{S}_i} = \dfrac{\text{Var}(\text{Var}_{X_i}(E_{X_{\sim i}}[\mathcal{Y}(X,\omega)|X_i]))}{(E[\text{Var}(\mathcal{Y}(X,\omega))])^2}
\end{equation}
where $\mu_{\tilde{S}_i}$ provides an estimate of the Sobol index and $\sigma^2_{\tilde{S}_i}$ gives an estimate of its uncertainty.
To compute the first-order Sobol indices, we start by taking the expectation of the GP $\mathcal{Y}(X,\omega)$ over all the inputs except `$X_i$' in the numerator of Eq.\ \eqref{eqn:Sobol_RV} and denote this as
\begin{equation}
    A_i(X_i,\omega) = E_{X_{\sim i}}[\mathcal{Y}(X,\omega)|X_i]
    \label{eq:marginal}
\end{equation}
Since, $\mathcal{Y}(X,\omega)$ is a Gaussian process and expectation is a linear operator, $A_i(X_i,\omega)$ is also a GP referred to as the main effect GP for $i$\textsuperscript{th} input dimension.

The mean and covariance function of the main effect GP can be determined by integrating the original GP with respect to the joint probability measure over all inputs except $X_i$. Considering independent inputs, the mean function is given by \cite{Oakley}: 
\begin{align}
    \mu_{A_i}(x_i) = E[A_i(X_i)] = \int_{\bm{x}_{\sim i}} \hat{y}(\bm{x})\prod_{j\neq i} p_{X_j}(x_j) dx_j 
    \label{eq:marginal_mean}
\end{align}
which is easily computed as a product of one-dimensional integrals. Again considering independent inputs, the covariance function of the main effect GP is given by \cite{Oakley}:
\begin{align}
    \text{Cov}(A_i(X_{1i}), A_i(X_{2i})) & = \int_{\bm{x}_{1\sim i}} \int_{\bm{x}_{2\sim i}} \text{Cov}(\mathcal{Y}(\bm{x}_1), \mathcal{Y}(\bm{x}_2)) \prod_{j\neq i} p_{X_j}(x_{1j}) d{x_{1j}} \prod_{k\neq i} p_{X_k}(x_{2k}) d{x_{2k}}
    \label{eq:marginal_cov}
\end{align}
which can be expressed as a product of 2-dimensional integrals. Detailed derivations for computing the mean and covariance functions of the main effect GPs in closed form under specific conditions are provided in Appendix \ref{app:1}. We further note that Eqs.\ \eqref{eq:marginal_mean} and \eqref{eq:marginal_cov} can be generalized for interactions between variables as shown in Appendix \ref{app:2}. Computation of the interaction sensitivities follows directly from the following for main effect sensitivities. These interaction sensitivities can, in principle, be used within the learning although this is outside the scope of the present work.
 
To compute Sobol indices, we need to compute the variance of the main effect GP as
\begin{equation}
\label{eq:qoi}
    \sigma^2_{A_i} = \int_{X_i} (A_i(x_i)-E[A_i(X_i)])^2 p_{X_i}(x_i) dx_i
\end{equation}
Here, a Monte Carlo estimate of this integral is computed by defining a random discretization of the sample space of $X_i$ as $\{a_1, a_2, \dots, a_{n_j}\}$ (note these points will be reused later for learning function evaluations as well) and constructing the following Gaussian random vector to approximate the main effect GP: 
\begin{align*}
V_i = [A_i(a_1), A_i(a_2), \hdots, A_i(a_{n_j})]^T
\end{align*}
We can then develop the following two Monte Carlo estimators of the Sobol indices.

\subsection{SI computation using the mean of the Main Effect GP}

In the first approach, the mean of vector $V_i$ (i.e. $\mu_{V_i} = E_{X_i}[V_i]$) is computed using Eqs.\ \eqref{eq:marginal_mean}, where each element of the mean vector is defined as $\mu_{V_{i, j}} = \mu_{A_i}(a_j) = E[A_i(a_j)]$ $\forall$ $j \in \{1, 2, \hdots, n_j \}$. Then, the following standard Monte Carlo estimators are used to estimate the variance of the main effect GP
\begin{equation}
    \hat{\sigma}_{A_i}^2 = \dfrac{1}{n_j-1} \sum_{j=1}^{n_j} (\mu_{V_{i, j}} - \Bar{\mu}_{V_i})^2
\end{equation}
where $\Bar{\mu}_{V_i} = \dfrac{1}{n_j} \sum_{j=1}^{n_j}\mu_{V_{i, j}} $ is the average of the mean vector. 
The first-order Sobol indices are then estimated by:
\begin{align}
    \mu_{S_i} = \frac{\hat{\sigma}_{A_i}^2}{\hat{\sigma}^2}  \quad \forall \quad i \in \{ 1, 2, \hdots, d\}
    \label{eqn:sobol_estimates1}
\end{align}
where $\hat{\sigma}^2$ is the total variance of $\mathcal{Y}(\bm{x})$. This estimator provides the Sobol index as identified from the mean GP predictor for the main effect, but does not account for uncertainty/variability in the main effect GP. In the next section, an estimator for the standard deviation of the Sobol index is also given.

\subsection{SI computation using the Main Effect GP}

Here, we simulate the main effect GP and use these simulations to estimate both the mean and the standard deviation of the Sobol index. First, the mean vector and covariance matrix of $V_i$ are computed using Eqs. \eqref{eq:marginal_mean} and \eqref{eq:marginal_cov}, respectively. We then generate $n_s$ realizations of the random vector $V_i$ using the Cholesky decomposition of the covariance matrix \citep{cressie2015statistics}, $C = L L^T$, where the $k^{th}$ realization for $i^{th}$ input (i.e. $V_{ik}$) is given as follows 
\begin{align}
    V_{ik} = E[A_i(X_i)] + L \epsilon_k = [A_{ik}(a_1), A_{ik}(a_2)), \hdots, A_{ik}(a_{n_j})]^T
    \label{eqn:main_effect_sim}
\end{align}
where $\epsilon_k$ is a realization of an $n_j$ dimensional zero-mean, uncorrelated Gaussian random vector. Then, the following standard Monte Carlo estimators are used to estimate the variance of the main effect GP
\begin{equation}
    \hat{\sigma}_{A_i}^2 = \dfrac{1}{n_s} \sum_{k=1}^{n_s} \hat{\sigma}_{A_{ik}}^2 = \dfrac{1}{n_s} \sum_{k=1}^{n_s} \dfrac{1}{n_j-1} \sum_{j=1}^{n_j} (A_{ik}(a_j) - \hat{\mu}_{V_{ik}})^2 
\end{equation}
where $\hat{\mu}_{V_{ik}}$ is the sample mean for vector $V_{ik}$, $\hat{\sigma}_{A_{ik}}^2$ is the $k^{th}$ estimate of the variance of main effect GP ($A_i$), and the corresponding standard error is given by
\begin{equation}
    \hat{s}_{A_i}^2 = \dfrac{1}{n_s-1} \sum_{k=1}^{n_s}(\hat{\sigma}_{A_{ik}}^2 - \hat{\sigma}_{A_i}^2 )^2.
\end{equation}
From these, the mean and standard deviation of the first-order Sobol indices are estimated as:
\begin{align}
    \mu_{S_i} = \frac{\hat{\sigma}_{A_i}^2}{\hat{\sigma}^2}  \quad \text{and} \quad
    \sigma^2_{S_i} = \frac{\hat{s}_{A_i}^2}{\hat{\sigma}^2} \quad \forall \quad i \in \{ 1, 2, \hdots, d\}
    \label{eqn:sobol_estimates2}
\end{align}
where $\hat{\sigma}^2$ is the total variance of  $\mathcal{Y}(\bm{x})$.

\subsection{Computational Considerations}

The computational cost of applying these two methods for Sobol index estimation are vastly different and can play a critical role when integrated into an active learning loop where estimates need to be computed repeatedly. Moreover, 
we have to repeat these estimates for each input dimension, which can be problematic for problems with a large number of inputs. This must be weighed against the accuracy of the two approaches and the need to estimate the standard deviation of the Sobol index. 
The first approach only utilizes the mean predictor of Main Effect GP, which is computationally inexpensive but doesn't estimate the uncertainty in Sobol indices. The second approach considers the complete probability structure of the Main Effect GP and therefore provides an error measure for each Sobol index. But it comes at a high computational cost because it requires a large set of sample points ($n_j$ should be large) to compute Sobol indices with high accuracy. This results in large covariance matrices, which can make the main effect simulations in Eq.\ \eqref{eqn:main_effect_sim} expensive. We therefore employ the first approach to estimate Sobol indices because we do not require uncertainty estimates on the Sobol indices at each iteration.

\section{Active Learning for Sobol Index Estimation}

Next, we compare strategies to perform active learning for Sobol Index estimation. We specifically consider two different learning functions designed with the GP-based Sobol index estimates from Eq.\ \eqref{eqn:Sobol_RV} in mind. We first recognize that the Sobol indices given by Eq.\ \eqref{eqn:Sobol_RV} are expressed as a ratio of two variances. The numerator is the variance of the main effect GP and can be expressed as $\sigma_{A_i}^2=\text{Var}(A_i(X_i))$, while the denominator is the expected value of the variance of the complete GP, $\mathcal{Y}(X,\omega)$. The challenge of active learning in this case is therefore to minimize uncertainty in both components. To do so is not straightforward. We therefore propose two learning strategies that aim to minimize uncertainty in either the numerator or the denominator.

According to the first strategy, we aim to minimize uncertainty in the overall GP surrogate (Sobol index denominator). This can be achieved using two existing, well-known learning functions, (a) Expected Improvement for Global Fit (EIGF - described in section \ref{sec:EIGF} -- which has been demonstrated to be highly effective for this task  \cite{lam2008sequential}) and (b) Variance Improvement for Global Fit (VIGF - described in section \ref{subsubsec: VIGF}). For the second strategy, we develop a new learning function, referred to as the MUSIC (Minimize Uncertainty in Sobol Index Computation) learning function, designed to select samples that reduce uncertainty in the main effect GPs $A_i(X_i)$ and described in Section \ref{sec:MUSIC}.  

\begin{figure}[!ht]
    \centering
    \includegraphics[width=0.9\textwidth]{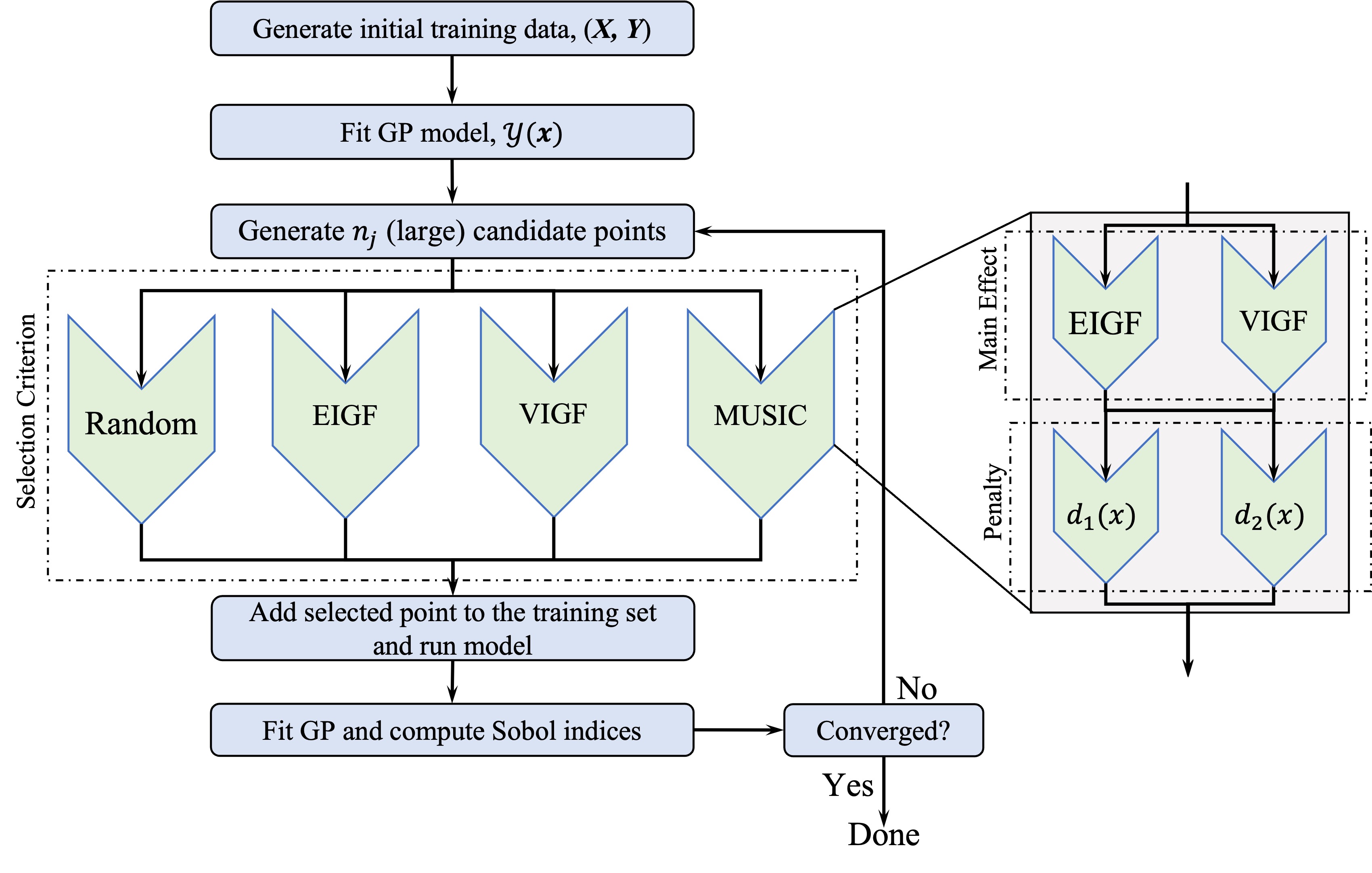}
    \caption{Flowchart of the active learning strategies for global sensitivity analysis. Each of the learning functions (and random sampling) shown in green are studied in this work.}
    \label{fig:flowchart}
\end{figure}

These two learning strategies are integrated into a standard active learning loop illustrated in Figure \ref{fig:flowchart}, which operates as follows. First, a set of initial training data $(\bm{X},\bm{Y})$ are generated by randomly sampling the input random vector $\bm{x}$ and evaluating the computational model $y(\bm{x})$ for each sample. Then, an initial GP surrogate model is fit to the data and initial estimates of the Sobol indices can be made using the estimators above. Next, we generate a large set of $n_j$ candidate samples of the input random variables (typically $n_j\ge 10000$) and evaluate the appropriate learning function at each candidate sample. According to the specified learning function, a candidate sample, say $\bm{x}^*$, is selected and the model is evaluated at this point, $y(\bm{x}^*)$. A new GP is fit to the augmented data set and the Sobol indices are updated. A convergence criterion on the Sobol indices is assessed. If the Sobol indices are deemed sufficiently accurate, the process stops. If the indices are not considered converged, then a new set of candidate points are generated and the iterations continue.

We compare both of these strategies with a simple random sampling approach in which, rather than employing a learning function in each iteration, one of the $n_j$ candidate samples is selected at random. This provides a baseline to measure whether the active learning strategies are effective for improving the convergence of Sobol index estimates.

\subsection{Expected Improvement for Global Fit (EIGF)}
\label{sec:EIGF}

To improve the fit of the global GP, $\mathcal{Y}(X,\omega))$, and specifically minimize uncertainty in the estimate of its variance, we employ the EIGF. Proposed by Lam \cite{lam2008sequential}, the EIGF defines an improvement function given by:
\begin{equation}
    I(\bm{x}) = (\mathcal{Y}(\bm{x}) - y(\bm{x}^{(j^*)}))^2 
    \label{eqn:improvement_function}
\end{equation}
where $y(\bm{x}^{(j^*)})$ is the observation of $y$ at the point $\bm{x}^{(j^*)}$ that is nearest to the point $\bm{x}$ and $\mathcal{Y}(\bm{x})$ is the normal random variable resulting from evaluating the GP at point $\bm{x}$. Taking the expectation of Eq.\ \eqref{eqn:improvement_function} yields
\begin{equation}
    \text{EIGF}(\bm{x}) = E[I(\bm{x})] = (\hat{y}(\bm{x})-y(\bm{x}^{(j^*)}))^2 + \sigma_{\hat{y}}^2(\bm{x})
    \label{eqn:EIGF}
\end{equation}
where, $\hat{y}(\bm{x})$ and $\sigma_{\hat{y}}^2(\bm{x})$ are the GP mean and variance at point $\bm{x}$ (see Eqs.\ \eqref{eqn:predictor} and \eqref{eqn:variance}). The EIGF learning function aims to identify the point $\bm{x}^*$ that maximizes the expectation in Eq.\ \eqref{eqn:EIGF} as: 
\begin{equation*}
    \bm{x}^* = \argmax_{\bm{x}}(\text{EIGF}(\bm{x}))
\end{equation*}
This point is selected as the new sample point for model evaluation and retraining of the GP surrogate model in the active learning loop shown in Figure \ref{fig:flowchart}.

A closer look at the EIGF in Eq.\ \eqref{eqn:EIGF} demonstrates a balance of two factors -- often referred to as \textit{exploration} and \textit{exploitation}. In the first term, $(\hat{y}(\bm{x})-y(\bm{x}^{(j^*)}))^2$, we see that the EIGF targets regions where variability in the function is large, which corresponds to exploitation. This is balanced by the second term, $\sigma_{\hat{y}}^2(\bm{x})$, which corresponds to regions where the GP prediction has high uncertainty, likely caused by a lack of training data in this region and encouraging exploration. However, it is well-known that the EIGF learning function favors exploitation (i.e.\ the first term often dominates) and may not explore new regions of parameter space sufficiently. Therefore a more balanced approach may be considered.

\subsection{Variance of Improvement for Global Fit (VIGF)}
\label{subsubsec: VIGF}

Mohammadi and Challenor \cite{mohammadi2022sequential} proposed a learning function using same improvement function as \textit{EIGF} function, given in Eq.~\eqref{eqn:improvement_function}.
They further recognized that $I(x)/\sigma^2(x)$ can be characterised by a non-central chi-square distribution (\cite{mohammadi2022sequential}).
\begin{equation}
    I(x)/\sigma^2(x) \sim \chi^{'2}\bigg( \kappa=1, \lambda=\bigg( \frac{\mu(x)-y(\bm{x}^{(j^*)})}{\sigma(x)}\bigg) \bigg)
\end{equation}
where the number of degree of freedom ($\kappa$) and noncentrality parameter ($\lambda$) are defined using the GP's posterior mean and standard deviation. Thus, the variance of the improvement function is expressed as
\begin{gather}
    \text{VIGF}(x) = Var\{I(x)\} = 4 \sigma^2(x) \left[(\mu(x) - y(\bm{x}^{(j^*)}))^2+2\sigma^2(x)\right].
\end{gather}

In the \textit{VIGF}, the first term ($4 \sigma^2(x)(\mu(x) - y(\bm{x}^{(j^*)}))^2$) is a product of terms that contribute toward local exploitation and global exploration, thus favoring both.  The second term ($8\sigma^4(x)$) focuses exclusively on global exploration. This learning function therefore gives more weight to global exploration than \textit{EIGF} and may provide a better balance in the exploration/exploitation trade off. 

\subsection{MUSIC Learning Function}
\label{sec:MUSIC}

From Eq.\ \eqref{eqn:sobol_estimates1} and \eqref{eqn:sobol_estimates2}, we see that the quality of the Sobol index estimates depends strongly on the quality of the main effect GPs. Large uncertainties in the variance of the main effect GPs will result in associated large uncertainties in Sobol index estimates. We therefore aim to create a training set that produces the most accurate main effect GPs. This problem, as we will see, can be posed in active learning terms as identifying the set of points that yields the best global fit for the main effect GPs, taking inspiration from the EIGF \cite{lam2008sequential} and VIGF \cite{mohammadi2022sequential}.

Let us first consider that we can construct an improvement function for the main effect GPs in each dimension, similar to Eq.\ \eqref{eqn:improvement_function}, as follows:
\begin{equation}
    I_i(x_i) = (A_i(x_i) - E_{X_{\sim i}}[Y|X_i](x^{(j^*)}_{i}))^2, \quad i \in\{1,\dots, d\} 
\end{equation}
where $x_{i}$, $x^{(j^*)}_{i}$ are the $i^{th}$ component of points $\bm{x}$ and $\bm{x}^{(j^*)}$, respectively and $x^{(j^*)}_{i}$ is the nearest training sample to $x_{i}$ in $i^{\text{th}}$ dimension. We immediately recognize that we cannot actually observe the true main effect $E_{X_{\sim i}}[Y(x^{(j^*)}_{i})|X_i]$. Instead, we must estimate this value using the main effect GP yielding a component-wise improvement function given by
\begin{equation}
    I_{A_i}(x_i) = (A_i(x_i) - A_i(x^{(j^*)}_{i}))^2, \quad i \in\{1,\dots, d\}
    \label{eqn:sobol_improvement}
\end{equation}
 Taking the expectation and variance of Eq.\ \eqref{eqn:sobol_improvement} yields:
\begin{equation}
    E[I_{A_i}(x_{i})] = (\mu_{A_i}(x_{i}) - \mu_{A_i}(x^{(j^*)}_{i}))^2 + \sigma^2_{A_i}(x_{i})
    \label{eqn:exp_sobol_improvement}
\end{equation}
\begin{equation}
    V[I_{A_i}(x_{i})] = 4\sigma^2_{A_i}(x_{i})[(\mu_{A_i}(x_{i}) - \mu_{A_i}(x^{(j^*)}_{i}))^2 + 2\sigma^2_{A_i}(x_{i})]
    \label{eqn:var_sobol_improvement}
\end{equation}
where $\mu_{A_i}(x_{i})$, $\mu_{A_i}(x^{(j^*)}_{i})$ can be computed using Eq.\ \eqref{eq:marginal_mean} and $\sigma^2_{A_i}(x_{i})$ can be computed from the diagonal terms of Eq.\ \eqref{eq:marginal_cov}. A closer look at $E[I_{A_i}(x_i)]$ demonstrates that it also balances exploration and exploitation specifically with regard to resolving the main effects. The first term targets regions where the main effects vary strongly with $X_i$ (exploitation) and the second term targets regions where the main effect GP has high uncertainty due to a lack of training data (exploration). Eq.~\eqref{eqn:var_sobol_improvement} has the same interpretation as the VIGF, balancing exploration and exploitation, but operating on the main effects.

Notice that the improvement criteria in Eq.~\eqref{eqn:exp_sobol_improvement} and \eqref{eqn:var_sobol_improvement} focus on improving the prediction of individual main effect GPs. These learning criteria can be collectively represented by a vector as follows:
\begin{align}
    \bm{M_A}(\bm{x}) = \begin{bmatrix} E[I_{A_1}(x_{1})] \\ E[I_{A_2}(x_{2})] \\ \vdots \\ E[I_{A_d}(x_{d}) \end{bmatrix} \quad \text{or} \quad \begin{bmatrix} V[I_{A_1}(x_{1})] \\ V[I_{A_2}(x_{2})] \\ \vdots \\ V[I_{A_d}(x_{d}) \end{bmatrix}
    \label{eq: music_main_effect}
\end{align}

From Eq.\ \eqref{eq: music_main_effect}, we could directly derive improvement-based learning functions over individual main effects. This might be useful, for example, if we were interested in resolving a specific term in the HDMR in Eq.\ \eqref{eqn:HDMR}. However, we are not interested only in individual main effects. Rather, we are interested in ensuring convergence of all main effects, to the extent possible. This means we must somehow balance the $d$ improvement functions that result from Eq.\ \eqref{eq: music_main_effect} to establish in single improvement function.  We do so by introducing a distance-based pre-factor that combines individual learning criteria on main effect GPs. This results in a new learning function, expressed as
\begin{align}
    \text{\Large$\eighthnote$}(\bm{x}) = \bm{D}^T(\bm{x}) \bm{M_A}(\bm{x})
\end{align} 
and termed the MUSIC (Minimizing Uncertainty in Sobol Index Convergence) learning function. We can consider numerous different distance pre-factors. Here, we specifically propose the following two pre-factors that produce convex combinations of main effect improvements
\begin{align}
   \bm{D}_1(\bm{x}) = \bm{W} \cdot \begin{bmatrix} |x_1 - x^{(j^*)}_{1}| \\ |x_2 - x^{(j^*)}_{2}|\\ \vdots \\ |x_d - x^{(j^*)}_{d}|\end{bmatrix}  \quad \text{and} \quad \bm{D}_2(\bm{x}) = \bm{W} \cdot ||\bm{x}- \bm{x}^{(j^*)}||_2^2 \bm{1}_d
\end{align}
where $||\bm{x}- \bm{x}^{(j^*)}||_2$ is the Euclidean distance between the sample $\bm{x}$ and nearest point $\bm{x}^{(j^*)}$, and $|x_i - x^{(j^*)}_{i}|$ is the distance between the $i^{\text{th}}$ component of $\bm{x}$ and nearest point $\bm{x}^{(j^*)}$ along $i^{\text{th}}$ dimension.  In either function, given two points whose total main effect improvements are comparable the algorithm should favor the point that is furthest from its nearest neighbor either in a Euclidean sense or in a component-wise sense. The weights ($\bm{W} = [w_1, w_2, \hdots, w_d]^T$) can be arbitrarily selected such that $\sum_i w_i=1$ to favor improvement in particular main effects. For example, these can be set equal to the existing Sobol index estimates in order to favor exploration in directions that correspond to higher Sobol indices. 

Using this learning function, we then select the next sample as the one that maximizes it as:
\begin{align*}
    \bm{x}^* = \argmax_{\bm{x}} \Large\eighthnote(\bm{x})
\end{align*}
from among the set of $n_j$ candidate samples. Table \ref{table: music_combinations} summarizes the learning function for four combinations based on the improvement criteria and the distance-based pre-factor.

\begin{table}[]
\centering
\begin{tabular}{|ll|l|}
\hline
\multicolumn{1}{|c|}{S.No.}  & \multicolumn{1}{|c|}{Combinations}                 &  \multicolumn{1}{c|}{Learning Function}\\ \hline
\multicolumn{1}{|l|}{1} & MUSIC+EIGF D1 &  $\Large\eighthnote_{E_1}(\bm{x}) = \sum_{i=1}^d w_i|x_i- x^{(j^*)}_i|E[I_{A_i}(x_{i})] $\\ \hline
\multicolumn{1}{|l|}{2} & MUSIC+EIGF D2 & $\Large\eighthnote_{E_2}(\bm{x}) = ||\bm{x}- \bm{x}^{(j^*)}||_2^2 \sum_{i=1}^d w_iE[I_{A_i}(x_{i})] $ \\ \hline
\multicolumn{1}{|l|}{3} & MUSIC+VIGF D1 &  $\Large\eighthnote_{V_1}(\bm{x}) = \sum_{i=1}^d w_i|x_i- x^{(j^*)}_i|V[I_{A_i}(x_{i})] $\\ \hline
\multicolumn{1}{|l|}{4} & MUSIC+VIGF D2 & $\Large\eighthnote_{V_2}(\bm{x}) = ||\bm{x}- \bm{x}^{(j^*)}||_2^2 \sum_{i=1}^d w_iV[I_{A_i}(x_{i})] $  \\ \hline
\end{tabular}
\caption{Four different combinations of MUSIC learning strategy based on improvement and distance-based pre-factor.}
\label{table: music_combinations}
\end{table}

\section{Exploration of Convergence using Analytical Functions}
\label{sec:results}

In this section, we study the convergence of the proposed active learning methods for four benchmark analytical functions with known Sobol indices and having increasing dimensions ranging from 2 to 15. Prior to presenting these examples, we provide a simple basis upon which to explain the observed convergence by considering the ratio of two quantities with some error. 

\subsection{Ratios of Errors}
The dependence on the absolute error of a ratio on estimates of the numerator and denominator is not straightforward. As we will see, reducing error in the numerator, denominator, or both does not necessarily reduce error in the ratio. Consider the following ratio:
\begin{align*}
    S = \frac{A}{Y}
\end{align*}
Next, consider that we have estimates of $S, A$ and $Y$ given by $\hat{S}, \hat{A}$ and $\hat{Y}$. 
Define the absolute error in the numerator ($A$) and denominator ($Y$) as:
\begin{align*}
    \Delta{A} = |A - \hat{A}| \\
    \Delta{Y} = |Y - \hat{Y}|
\end{align*}
such that  
\begin{align*}
    \hat{A} = A \pm \Delta{A} \\
    \hat{Y} = Y \pm \Delta{Y}
\end{align*}
Next, the absolute error in the ratio can be expressed as:
\begin{align*}
    \Delta{S} = |S - \hat{S}| =  \Bigg| S -\frac{\hat{A}}{\hat{Y}} \Bigg|\\
    \Delta{S} = \Bigg|S - \frac{A \pm \Delta{A}}{Y\pm\Delta{Y}}\Bigg| \\
    \Delta{S} = \Bigg| \frac{S (Y\pm\Delta{Y}) - A \mp \Delta{A}}{Y\pm\Delta{Y}} \Bigg| \\
    \Delta{S} = \Bigg| \frac{SY \pm S\Delta{Y} - A \mp \Delta{A}}{Y\pm\Delta{Y}} \Bigg| \\
    \Delta{S} = \Bigg| \frac{\pm S\Delta{Y} \mp \Delta{A}}{Y\pm\Delta{Y}} \Bigg|,
\end{align*}
which depends, in a nontrivial way, on the errors $\Delta A$ and $\Delta Y$ -- but also depends on the value of $S$ itself. 

Recognizing that our GP models for $A$ and $Y$ presented in the formulation of the GP-based Sobol indices above may be systematically biased (depending on the training set and learning function being used) and that we may not know this bias a priori, we explore this error under four specific cases. 

\begin{itemize}
    \item Case 1: $\hat{A}$ and $\hat{Y}$ are both underestimated such that $\hat{A} = A - \Delta{A}$ and $\hat{Y} = Y - \Delta{Y}$ yielding
    \begin{align*}
    \Delta{S} =  \frac{|- S\Delta{Y} + \Delta{A}|}{Y - \Delta{Y}}
    \end{align*}

    \item Case 2: $\hat{A}$ and $\hat{Y}$ are both overestimated such that $\hat{A} = A + \Delta{A}$ and $\hat{Y} = Y + \Delta{Y}$ yielding
    \begin{align*}
    \Delta{S} =  \frac{| S\Delta{Y} - \Delta{A}|}{Y + \Delta{Y}}
    \end{align*}
    
    \item Case 3: $\hat{A}$ is overestimated ($\hat{A} = A + \Delta{A}$) and $\hat{Y}$ is underestimated ($\hat{Y} = Y - \Delta{Y}$) yielding
    \begin{align*}
    \Delta{S} =  \frac{| -S\Delta{Y} - \Delta{A}|}{Y - \Delta{Y}}
    \end{align*}
    
    \item Case 4: $\hat{A}$ is underestimated ($\hat{A} = A - \Delta{A}$) and $\hat{Y}$ is overestimated ($\hat{Y} = Y + \Delta{Y}$) yielding
    \begin{align*}
    \Delta{S} =  \frac{| S\Delta{Y} + \Delta{A}|}{Y +\Delta{Y}}
    \end{align*}

\end{itemize}

\begin{figure}[!ht]
    \centering
    \includegraphics[width=\textwidth]{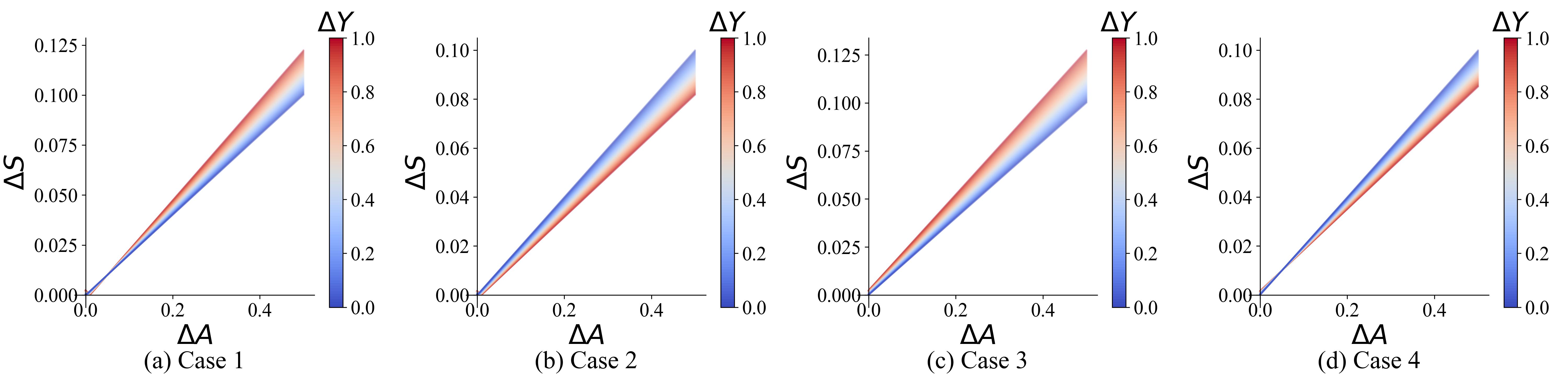}
    \caption{Influence of errors in the numerator, $\Delta{A}$, and denominator, $\Delta{Y}$, on error $\Delta{S}$ in the ratio $S=A/Y$ for $S=0.01$ for four cases of systematic bias (a) Case 1: Both $\Delta{A}$ and $\Delta{Y}$ are underestimated, (b) Case 2: Both $\Delta{A}$ and $\Delta{Y}$ are overestimated, (c) $\Delta{A}$ is overestimated and $\Delta{Y}$ is underestimated, and (d) $\Delta{A}$ is underestimated and $\Delta{Y}$ is overestimated.}
    \label{fig:err_case1}
\end{figure}

\begin{figure}[!ht]
    \centering
    \includegraphics[width=\textwidth]{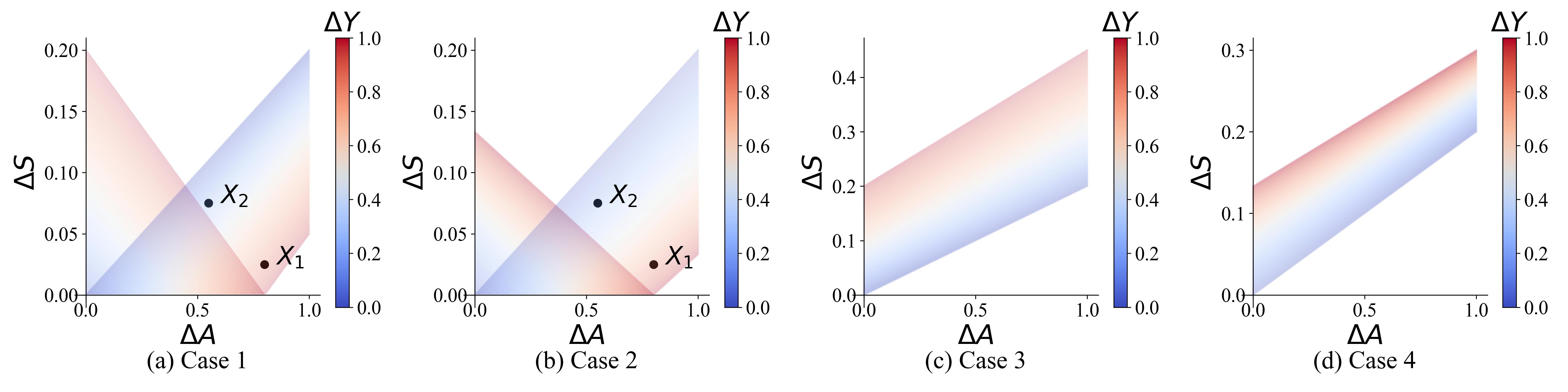}
    \caption{Influence of errors in the numerator, $\Delta{A}$, and denominator, $\Delta{Y}$, on error $\Delta{S}$ in the ratio $S=A/Y$ for $S=0.8$ for four cases of systematic bias (a) Case 1: Both $\Delta{A}$ and $\Delta{Y}$ are underestimated, (b) Case 2: Both $\Delta{A}$ and $\Delta{Y}$ are overestimated, (c) $\Delta{A}$ is overestimated and $\Delta{Y}$ is underestimated, and (d) $\Delta{A}$ is underestimated and $\Delta{Y}$ is overestimated.}
    \label{fig:err_case2}
\end{figure}

Figures \ref{fig:err_case1} and \ref{fig:err_case2} show plots of the dependence of $\Delta S$ on $\Delta A$ and $\Delta Y$ for $S=0.01$ and $S=0.8$, respectively, in each of these four cases for a hypothetical ratio. Notice that, for $S=0.01$ (Figure \ref{fig:err_case1}) the errors appear to decrease toward zero as both $\Delta A$ and $\Delta Y$ decrease \footnote{In fact, Cases 1 and 2 have the same bias observed in the $S=0.8$ case but the influence is negligible because $S$ is very small.}. That is, if either (or both) $\Delta A$ and $\Delta Y$ decrease, then $\Delta S$ decreases. However, for $S=0.8$, a decrease in either $\Delta A$ or $\Delta Y$ (or both) does not guarantee a decrease in $\Delta S$. In Cases 1 and 2, regardless of $\Delta A$ (even for very high $\Delta A$), it is possible to obtain zero $\Delta S$. In fact, for a fixed value of $\Delta Y$, as $\Delta A$ decreases $\Delta S$ will decrease to zero and then begin increasing again. This occurs because the systematic bias in both Case 1 and Case 2 can cause the ratio $S=A/Y$ to be correct even if both $A$ and $Y$ are (perhaps severely) incorrect. Indeed, two points $X_1$ and $X_2$ are shown in Figure Y where $\Delta A_{X_1}>\Delta A_{X_1}$ and $\Delta Y_{X_1}>\Delta Y_{X_2}$, yet $\Delta S_{X_1} < \Delta S_{X_2}$. As we will see, this systematic bias in $A$ and $Y$ can make accelerated convergence in the Sobol indices difficult. 

Next, we will study the convergence, and specifically the behavior of errors in light of the discussion above, for the different active learning schemes using four analytical test cases of increasing dimension.

\subsection{Square Exponential Function}
\label{sec:exponential}
\begin{figure}[!ht]
    \centering
    \includegraphics[ width=0.9\textwidth]{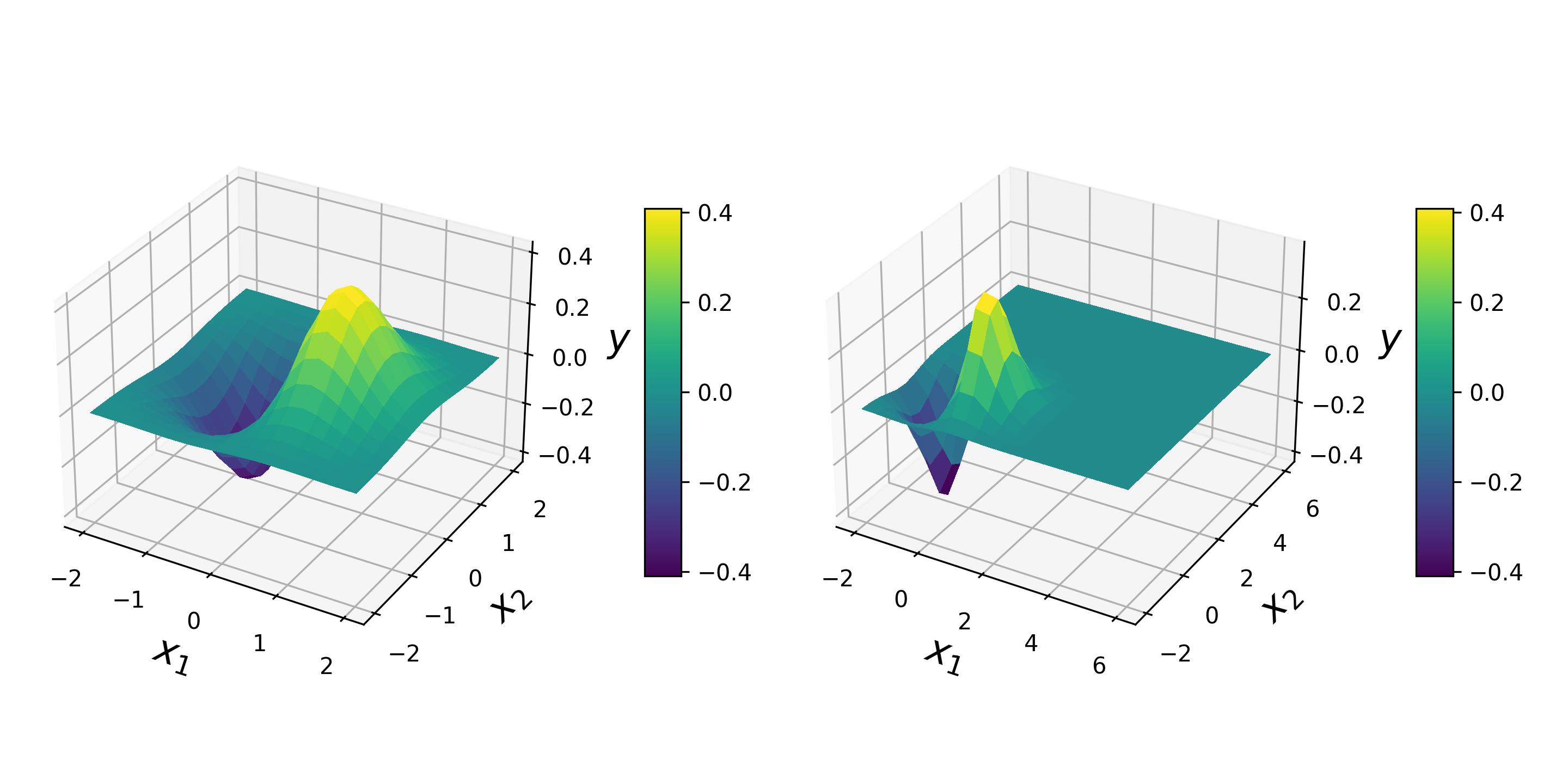}
    \caption{Square exponential functions (a) with tight bounds and no flat region and (b) with wide bounds and large flat region.}
    \label{Exp_fun}
\end{figure}

First, we consider the following square exponential function of two random inputs (i.e. $\mathbf{X} = [X_1, X_2]$).
\begin{align*}
    y = X_1 \exp{(-X_1^2-X_2^2)}
\end{align*}
where, $X_1$ and $X_2$ are uniformly distributed between [a, b]. Two cases are considered in this study, one with tight bounds around the two peaks ($a=-2$ and $b=2$) and another with larger bounds $b$ ($a=-2$ and $b=6$) such that a large flat region is observed across much of the domain. Both cases are illustrated in Figure \ref{Exp_fun}. The presence of this flat region has a significant impact on the Sobol indices and, moreover, affects the convergence of the active learning algorithms. In both cases, the Sobol indices can be computed analytically. For the first case ($b=2$), the first-order Sobol indices are $S_1 = 0.6208$ and $S_2=0$. For the second case, the  Sobol indices are $S_1 = 0.3119$ and $S_2=2.30 \times 10^{-5}$.
\begin{figure}[!ht]
    \centering
    \includegraphics[width=0.9\textwidth]{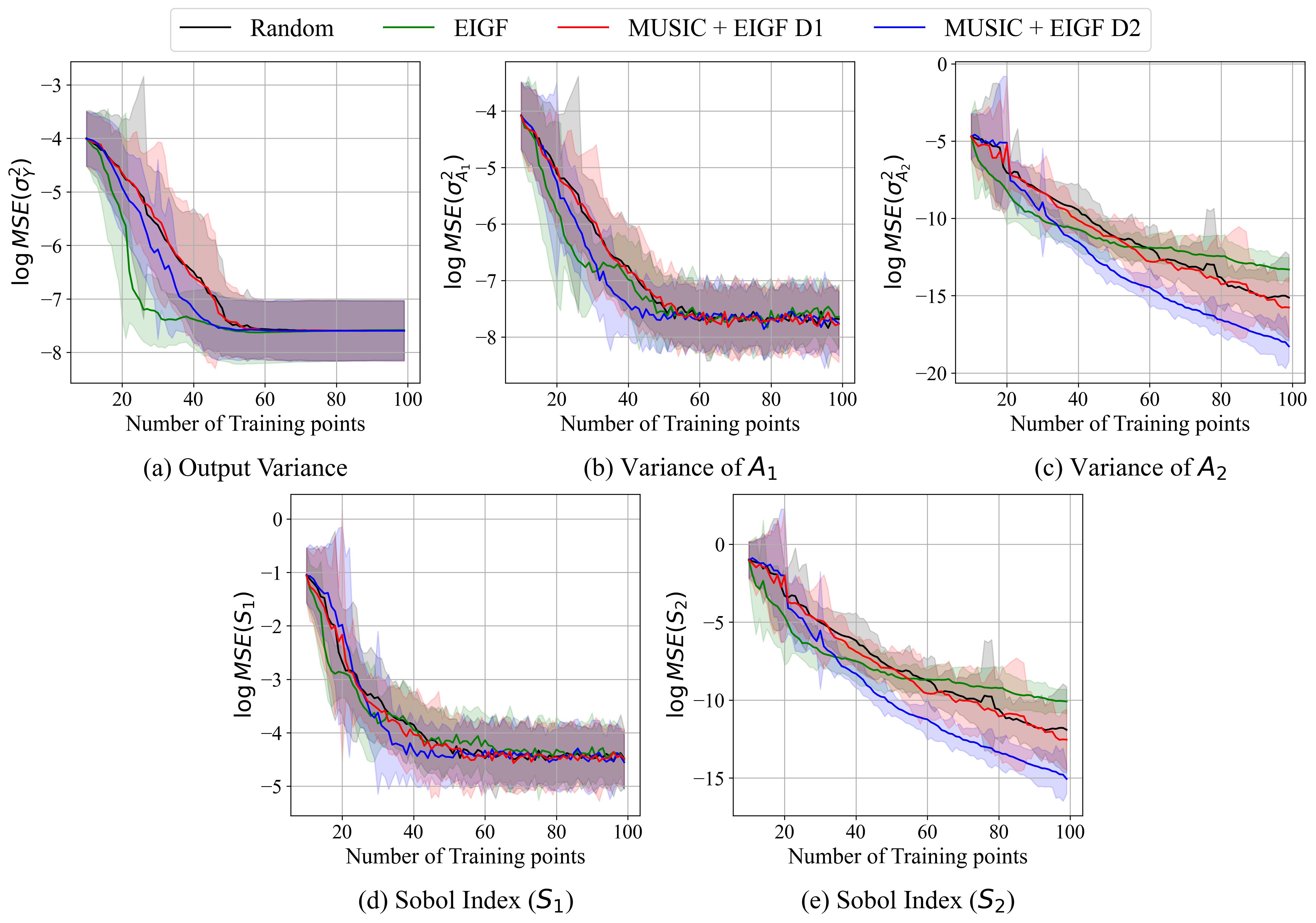}
    \caption{Exponential Function ($b=2$, no flat region) -- Mean square convergence with $\sigma$ confidence intervals from 100 repeated trials of (a) output variance $\sigma_Y^2$, (b,c) main effect variances $\sigma_{A_1}^2$ and $\sigma_{A_2}^2$, and (d,e) Sobol indices $S_1$ and $S_2$ for MUSIC and EIGF active learning schemes compared with random sampling.}
    \label{exp_b2_eigf}
\end{figure}
\begin{figure}[!ht]
    \centering
    \includegraphics[width=0.9\textwidth]{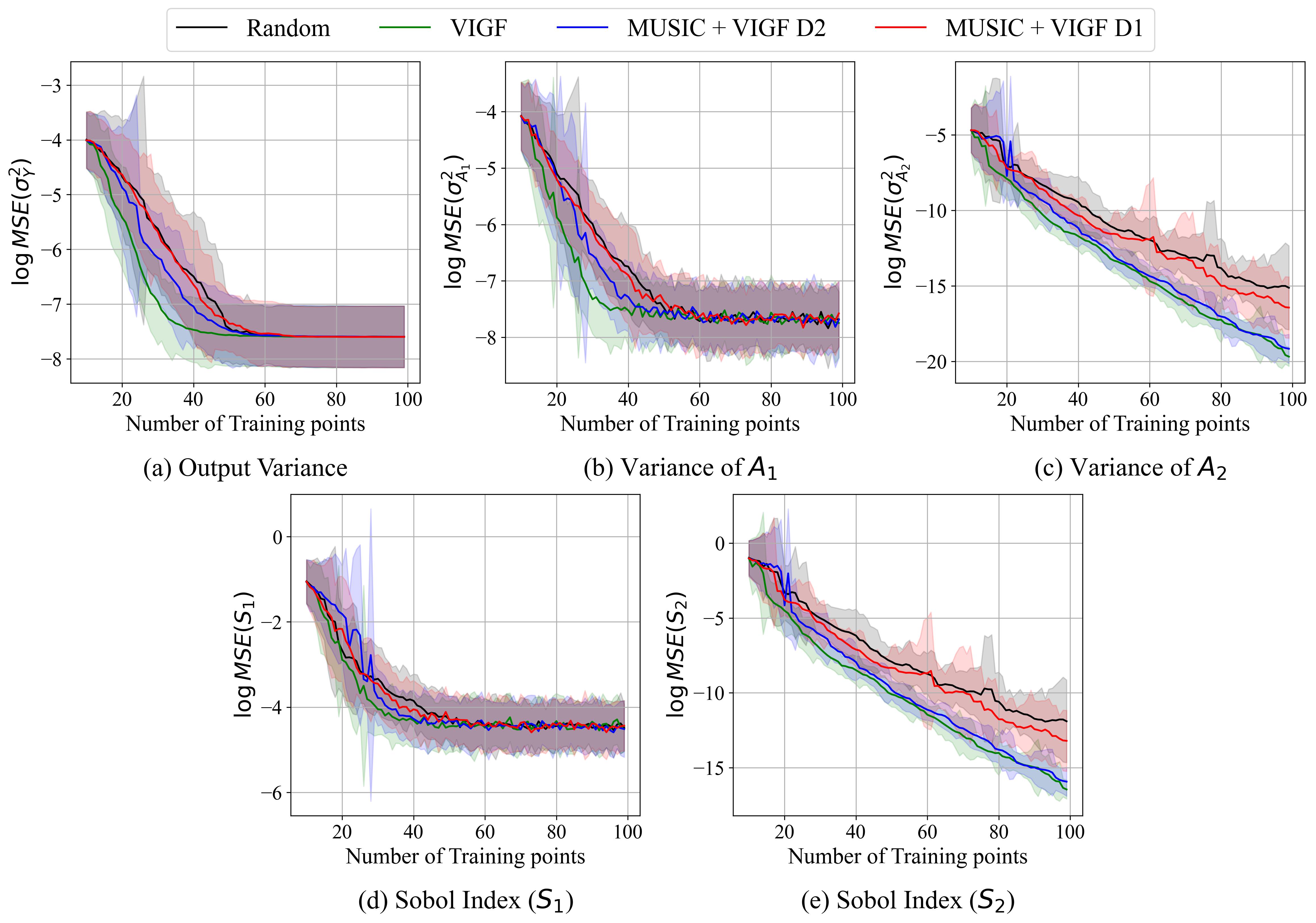}
    \caption{Exponential Function ($b=2$, no flat region) -- Mean square convergence with $\sigma$ confidence intervals from 100 repeated trials of (a) output variance $\sigma_Y^2$, (b,c) main effect variances $\sigma_{A_1}^2$ and $\sigma_{A_2}^2$, and (d,e) Sobol indices $S_1$ and $S_2$ for MUSIC and VIGF active learning schemes compared with random sampling.}
    \label{exp_b2_vigf}
\end{figure}

Figure \ref{exp_b2_eigf} shows the mean square error convergence from 100 repeated trials of the different active learning schemes using the EIGF and MUSIC learning functions compared with random sampling for the case where $b=2$ (no flat region). Figure~\ref{exp_b2_eigf}(a) shows convergence of the variance of the output, $\sigma^2_Y$, as estimated by the GP. We see that the EIGF generally outperforms the other two approaches, converging to a very accurate estimate for $\sigma^2_Y$ with less than $\sim 30$ samples. Figures~\ref{exp_b2_eigf}(b) and (c) show the same convergence plots for the variance of the main effect GPs $\sigma_{A_1}^2$ and $\sigma_{A_2}^2$, respectively. We see that, in main effects the MUSIC learning function offers comparable convergence to the the EIGF in $A_1$ and superior convergence in $A_2$. Both methods generally outperform random sampling in the main effect variances. 

Finally, figures~\ref{exp_b2_eigf}(d) and (e) show convergence of the Sobol indices $S_1$ and $S_2$. For this problem, the Sobol convergence tracks closely with the main effect convergence, but with slight differences. For example, we recall that the variance $\sigma_{A_1}^2$ from Figure~\ref{exp_b2_eigf}(b) is the numerator and $\sigma_Y^2$ from Figure~\ref{exp_b2_eigf}(a) is the denominator of the Sobol index $S_1$. However, despite superior performance in both $\sigma_{A_1}^2$ and $\sigma_{Y}^2$, the MUSIC function performs mildly worse than random sampling in $S_1$ for small samples. This is due to the `ratio of errors' issues discussed in the previous section.

Figure~\ref{exp_b2_vigf} shows the same convergence plots but compares VIGF and MUSIC (based on VIGF) learning functions against the sequential random sampling. Similar to EIGF, the VIGF learning function results in the best estimate for output variance ($\sigma_Y^2$), as shown in Fig. \ref{exp_b2_vigf}(a). However, VIGF outperforms EIGF for main effect and sobol estimates ($\sigma_{A_i}^2$) and yields similar convergence to the `MUSIC+VIGF D2' combination. This improvement is due to the better prediction at both functional features (peak and valley). Consider, the main effect of first dimension $E[Y|X_1]$ for a fixed value of $X_1$ (imagine a cross-section at arbitrary $X_1$ in Figure~\ref{Exp_fun}), the main effect function depends on either the peak or the valley (depending on the value of $X_1$). But, the main effect of the second dimension ($E[Y|X_2]$) depends on both features for all values of $X_2$. Since VIGF gives more importance to exploration (as compared to EIGF), it resolves both features simultaneously -- it does not get stuck exploiting one feature. This results in better estimates for $\sigma_Y^2$ as well as $\sigma_{A_i}^2$ and eventually better Sobol estimates as shown in \ref{exp_b2_vigf}(e).

\begin{figure}[!ht]
    \centering
    \includegraphics[width=0.9\textwidth]{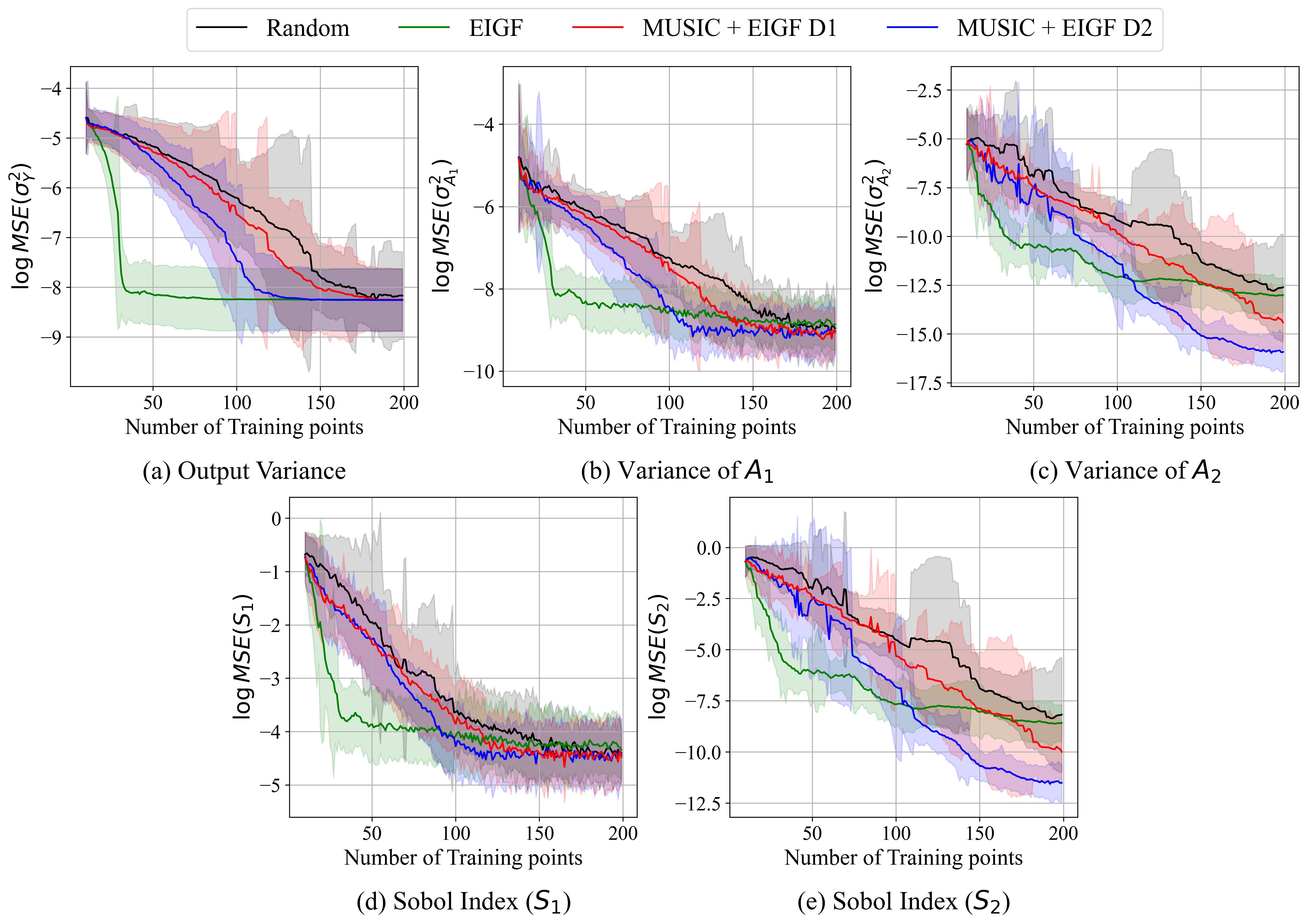}
    \caption{Exponential Function ($b=6$, with flat region) -- Mean square convergence with $\sigma$ confidence intervals from 100 repeated trials of (a) output variance $\sigma_Y^2$, (b,c) main effect variances $\sigma_{A_1}^2$ and $\sigma_{A_2}^2$, and (d,e) Sobol indices $S_1$ and $S_2$ for MUSIC and EIGF active learning schemes compared with random sampling.}
    \label{exp_b6_eigf}
\end{figure}

\begin{figure}[!ht]
    \centering
    \includegraphics[width=0.9\textwidth]{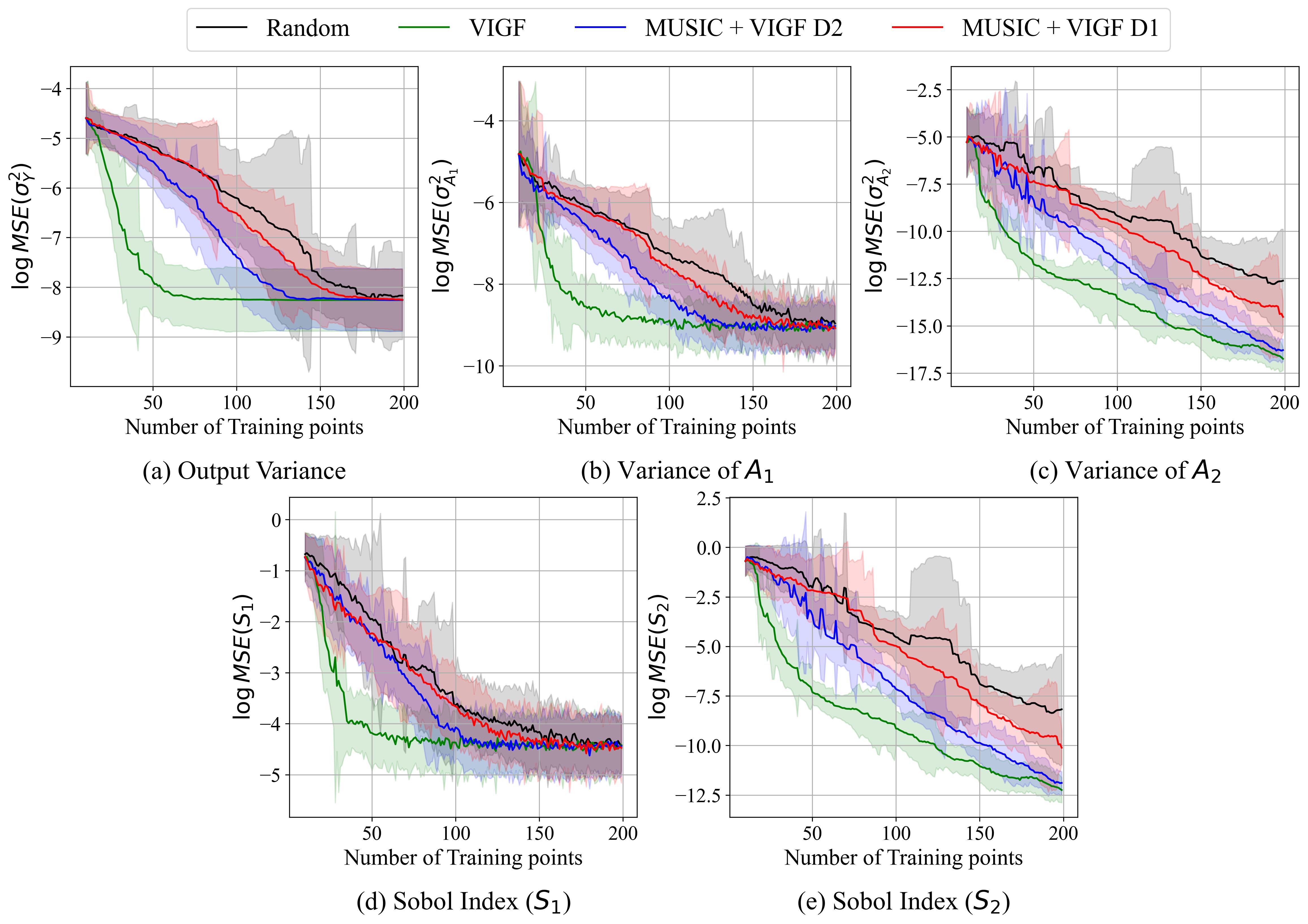}
    \caption{Exponential Function ($b=6$, with flat region) -- Mean square convergence with $\sigma$ confidence intervals from 100 repeated trials of (a) output variance $\sigma_Y^2$, (b,c) main effect variances $\sigma_{A_1}^2$ and $\sigma_{A_2}^2$, and (d,e) Sobol indices $S_1$ and $S_2$ for MUSIC and VIGF active learning schemes compared with random sampling.}
    \label{exp_b6_vigf}
\end{figure}

Figure~\ref{exp_b6_eigf} and \ref{exp_b6_vigf} show the same convergence plots for the case with $b=6$ (having a flat region). As expected, the EIGF/VIGF learning functions perform much better than the other methods for $\sigma_Y^2$, as they both perform well for tasks where exploitation is paramount \cite{beck2016sequential}. They both also exhibit superior performance in the main effect variances $\sigma_{A_1}^2$ and $\sigma_{A_2}^2$ and Sobol indices $S_1$ and $S_2$ for small sample sizes. Again, this is a result of its strong exploitation properties. The MUSIC function meanwhile provides superior convergence to random sampling in all cases as well. 

\subsection{Ishigami function}
Next, consider the well-known Ishigmai function with three independent input random variables ($\mathbf{X} = [X_1, X_2, X_3]$) that are uniformly distributed over [$-\pi, \pi$], and defined by:
\begin{align*}
y(\mathbf{x}) = \sin{x_1} + a\sin^2{x_2} + bx_3^4 \sin{x_1}
\end{align*}
where parameters $a=7$, and $b=0.1$. This function has nonlinear and nonmonotonic behavior, but the third input dimension exhibits a peculiar behavior. The first-order contribution of the third dimension is zero towards output variance, but it can't be completely ignored as an interaction sensitivity index (i.e. $S_{13}$) has a significant impact on output variance. The first-order Sobol indices are $S_1=0.3139$, $S_2=0.4424$, and $S_3=0$, determined analytically by solving for the variance of the main effect and output as 
\begin{align*}
    Var\{A_1\} = \frac{1}{2}\Big(1+\frac{b \pi^4}{5}\Big)^2 , \quad Var\{A_2\} = \frac{a^2}{8}, \quad Var\{A_3\} = 0
\end{align*}
\begin{equation*}
    Var\{Y\} = \frac{a^2}{8} + \frac{b \pi^4}{5} + \frac{b^2 \pi^8}{18} + \frac{1}{2}
\end{equation*}

\begin{figure}[!ht]
    \centering
    \includegraphics[width=\textwidth]{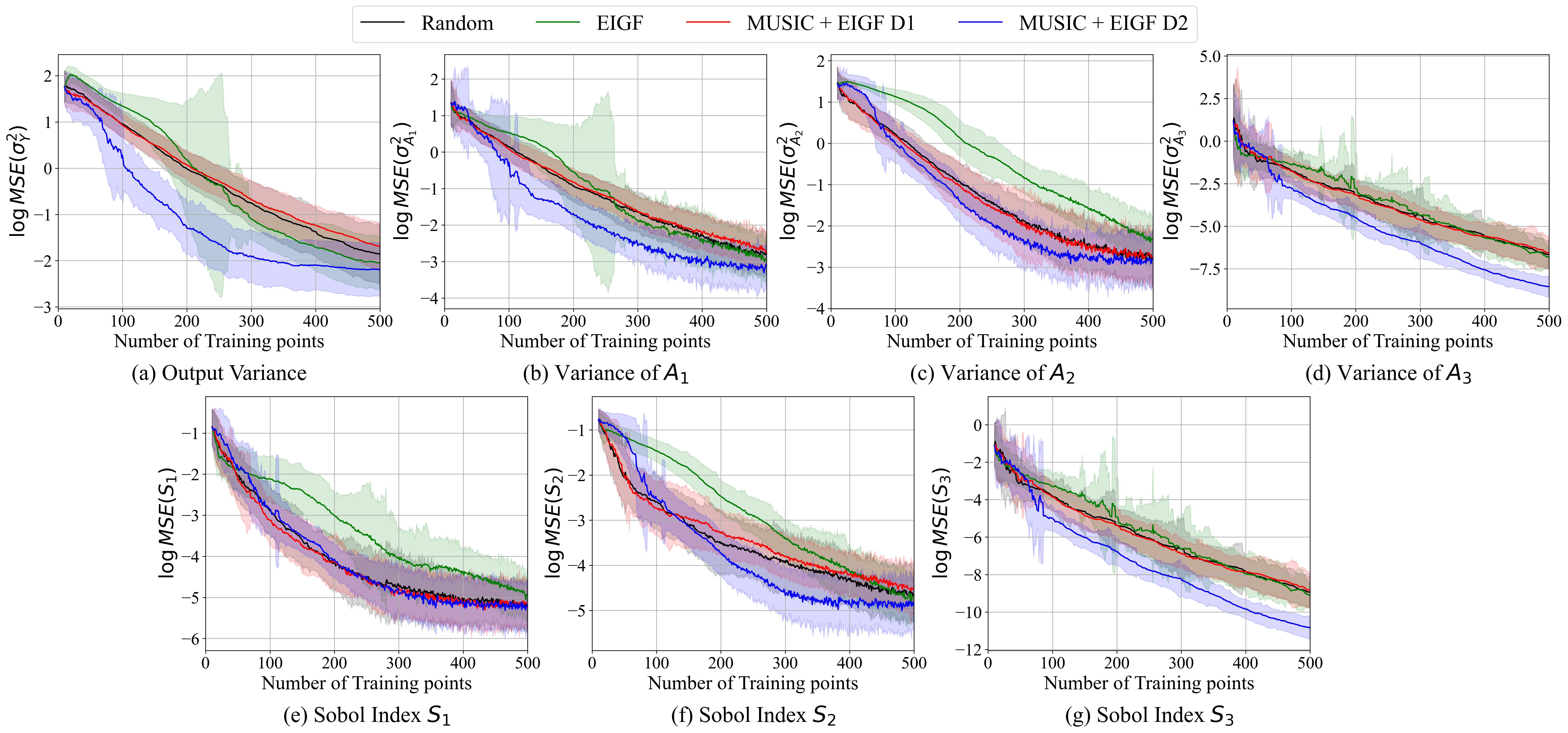}
    \caption{Ishigami Function -- Mean square convergence with $\sigma$ confidence intervals from 100 repeated trials of (a) output variance $\sigma_Y^2$, (b-d) main effect variances $\sigma_{A_i}^2$ $\forall$ $i \in \{1, 2, 3\}$, and (e-g) Sobol indices $S_1$, $S_2$, and $S_3$ for MUSIC and EIGF active learning schemes compared with random sampling.}
    \label{ish_si_eigf}
\end{figure}

The performance of EIGF and MUSIC (`EIGF D1' and `EIGF D2') adaptive sampling strategies is shown in Figure. \ref{ish_si_eigf}. The adaptive algorithms start with a small set of 10 samples generated using Latin Hypercube Sampling and convergence is observed up to 500 samples. After updating the surrogate model, Sobol indices are estimated using 25,000 candidate points. The process is repeated for 100 trials to obtain confidence intervals in the convergence. 
In this example, the EIGF struggles to accurately estimate the output variance ($\sigma^2_Y$ in Figure~\ref{ish_si_eigf}(a)) and the main effect variances ($\sigma_{A_i}^2$ in Figure~\ref{ish_si_eigf}(b)--(d)).
This can be explained by the sinusoidal nature of the Ishigami function, which requires a balance between exploration and exploitation. 
Error convergence of the `MUSIC+EIGF D1' learning function is similar to random sampling for every estimate. `MUSIC+EIGF D2', on the other hand, leads to superior convergence in nearly all aspects. The distance-based pre-factor is the main difference here. The $D_2$ function uses the Euclidean distance and ensures that training samples are globally far away from each other, which helps the `MUSIC+EIGF' learning function to avoid local exploitation. Because the $D_1$ function utilizes an element-wise distance component, it does not avoid local exploitation as effectively. Figures~\ref{ish_si_eigf} (e-g) show the ultimate convergence of the Sobol index estimates where again `MUSIC+EIGF D2' generally outperforms the other criteria. 

\begin{figure}[!ht]
    \centering
    \includegraphics[width=\textwidth]{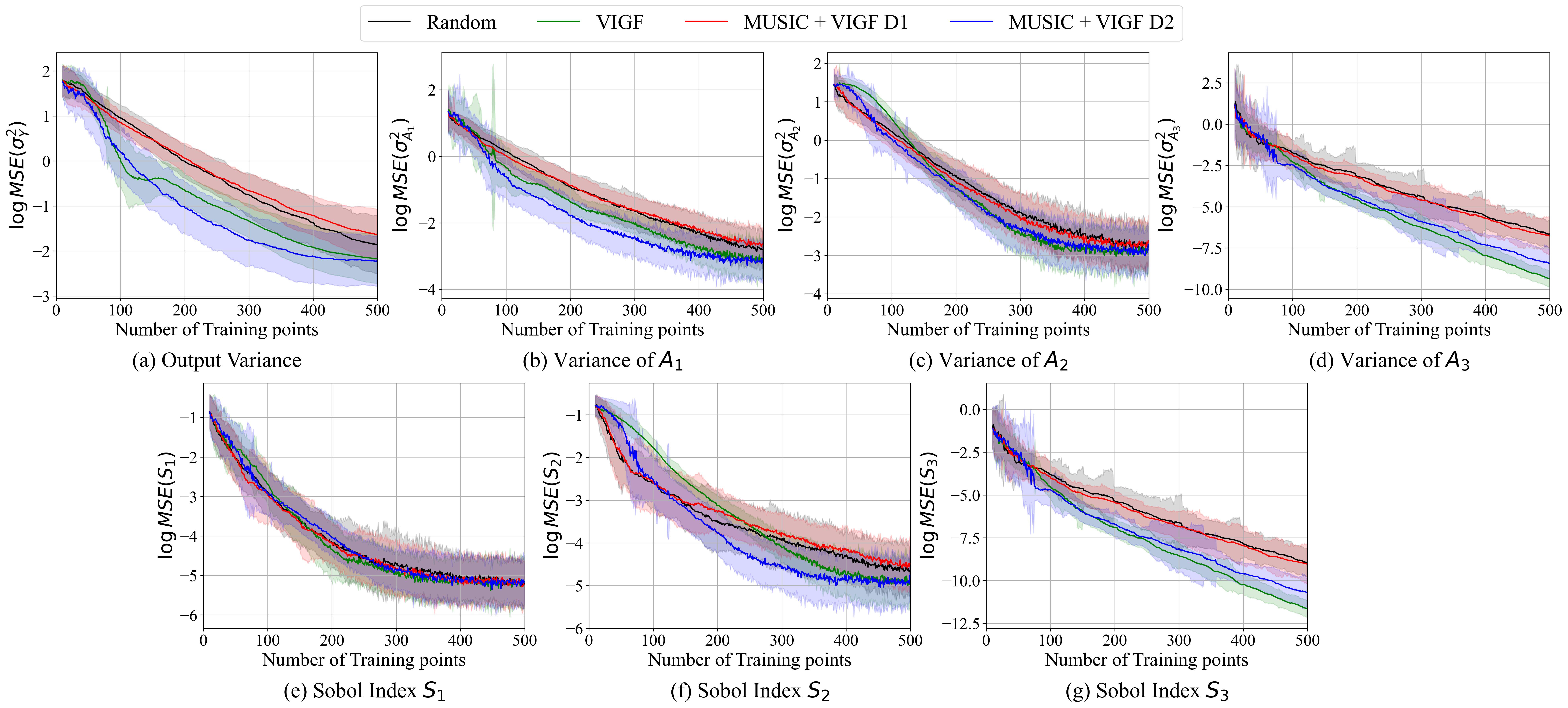}
    \caption{Ishigami Function -- Mean square convergence of (a) output variance $\sigma_Y^2$, (b-d) main effect variances $\sigma_{A_i}^2$ $\forall$ $i \in \{1, 2, 3\}$, and (e-g) Sobol indices $S_1$, $S_2$, and $S_3$ for MUSIC and VIGF active learning schemes compared with random sampling.}
    \label{ish_si_vigf}
\end{figure}

Figure \ref{ish_si_vigf} likewise compares the VIGF and MUSIC (+VIGF) learning functions to random sampling. The main difference here is that the VIGF improves performance considerably over the EIGF. Given this consistent improvement with the VIGF, we will compare only the results using VIGF in the remaining analytical examples.

\begin{figure}[!ht]
    \centering
    \includegraphics[trim={5cm 1cm 0 0.5cm},clip,width=1.05\textwidth]{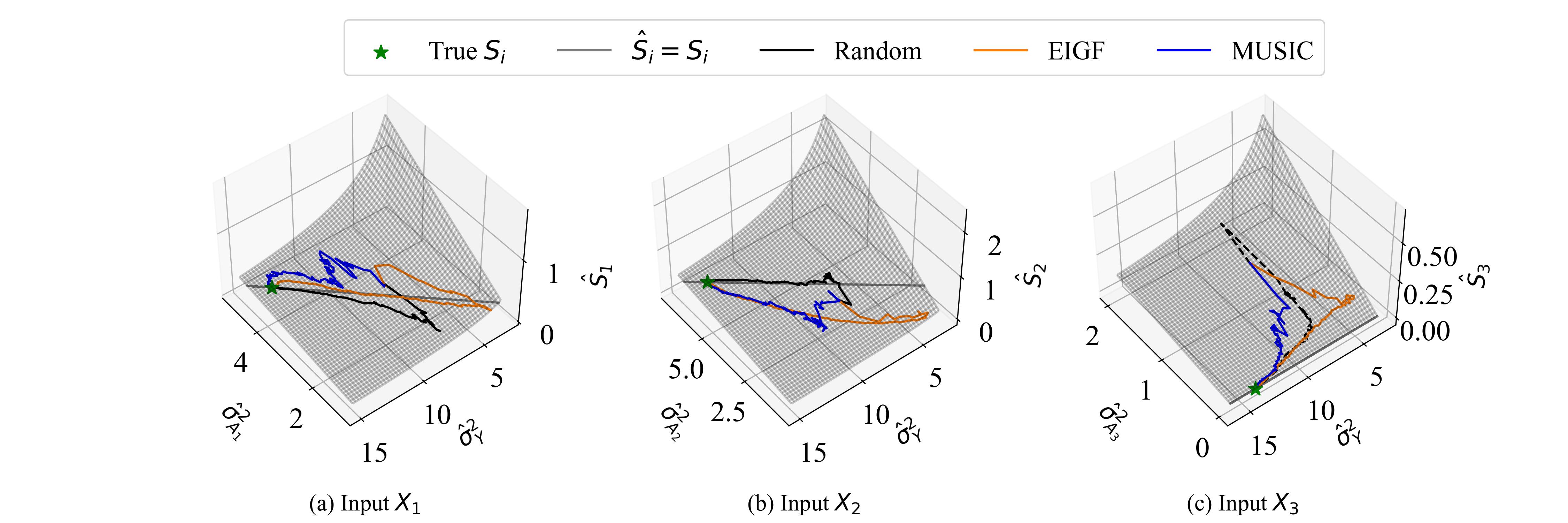}
    \caption{Dependence of Sobol indices on the variance of main effect GPs and Output GP.}
    \label{ratio}
\end{figure}

Here, the convergence of the first and second Sobol indices may appear counterintuitive. The MUSIC function estimates both the total and main effect variances with high accuracy, but the error in the Sobol estimates is comparable to random sampling. This is again due to the effect of ratios of errors. Figure. \ref{ratio} 
shows the convergence path of the Sobol indices by plotting traces of $(\hat{\sigma}^2_Y,\hat{\sigma}^2_{A_i})$ averaged over 100 trials.
The grey surface represents the Sobol index for any specific pair. The green star is the actual first-order Sobol index ($S_i$) value for the $i$\textsuperscript{th} dimension. The grey line shows all combinations of $(\hat{\sigma}^2_Y,\hat{\sigma}^2_{A_i})$, such that their ratio is equal to the true Sobol index (i.e. $\hat{S}_i = S_i$). The black, orange, and blue curves illustrate the convergence of random, EIGF, and MUSIC sampling, respectively. We notice that, although MUSIC and EIGF perform better in the variance estimators (see convergence above), the path that their estimators takes toward the true value does not follow the grey line. That is, their Sobol index estimates are biased due to the ratio of errors. This is not the case for random sampling, which converges largely along the gray line.
This counter-intuitive behavior is not observed for the third dimension because the actual Sobol index is zero (i.e. $S_3 = 0$) and therefore depends only on the main effect GP. As illustrated in Figure \ref{fig:err_case1}, when $S_i$ is very small convergence in the numerator will yield convergence in the ratio. We can therefore conclude that the MUSIC function will be very effective at identifying small Sobol indices.

\subsection{G-function}
Next, we consider the 5-dimensional G-function, defined as the following product of non-linear functions in each dimension, 
\begin{align*}
y(\mathbf{x}) = \prod_{k=1}^{d} \frac{|4x^{(k)}-2|+a^{(k)}}{a^{(k)}+1}
\end{align*}
where each input feature is uniformly distributed between [0, 1] (i.e. $X^{(k)} \sim \text{Uniform}(0, 1)$).  
The analytical Sobol indices in five dimensions (i.e. $d=5$) are 0.48, 0.21, 0.12, 0.08, 0.06, corresponding to $a^{(k)} = k \text{, } k \in {1, 2, 3, 4, 5}$ from the following equations 
\begin{align*}
    Var\{A_i\} = \frac{1}{3(1+a^{(i)})^2} \quad \quad Var\{Y\} = \Big[\prod_{k=1}^d (1+Var\{A_k\})\Big] - 1
\end{align*}
\begin{align*}
S_i = \frac{1/(3(1+a^{(i)})^2)}{\Big[\prod_{k=1}^d{[1+1/(3(1+a^{(k)})^2)}]\Big]-1}
\end{align*}

\begin{figure}[!ht]
    \centering
    \includegraphics[width=\textwidth]{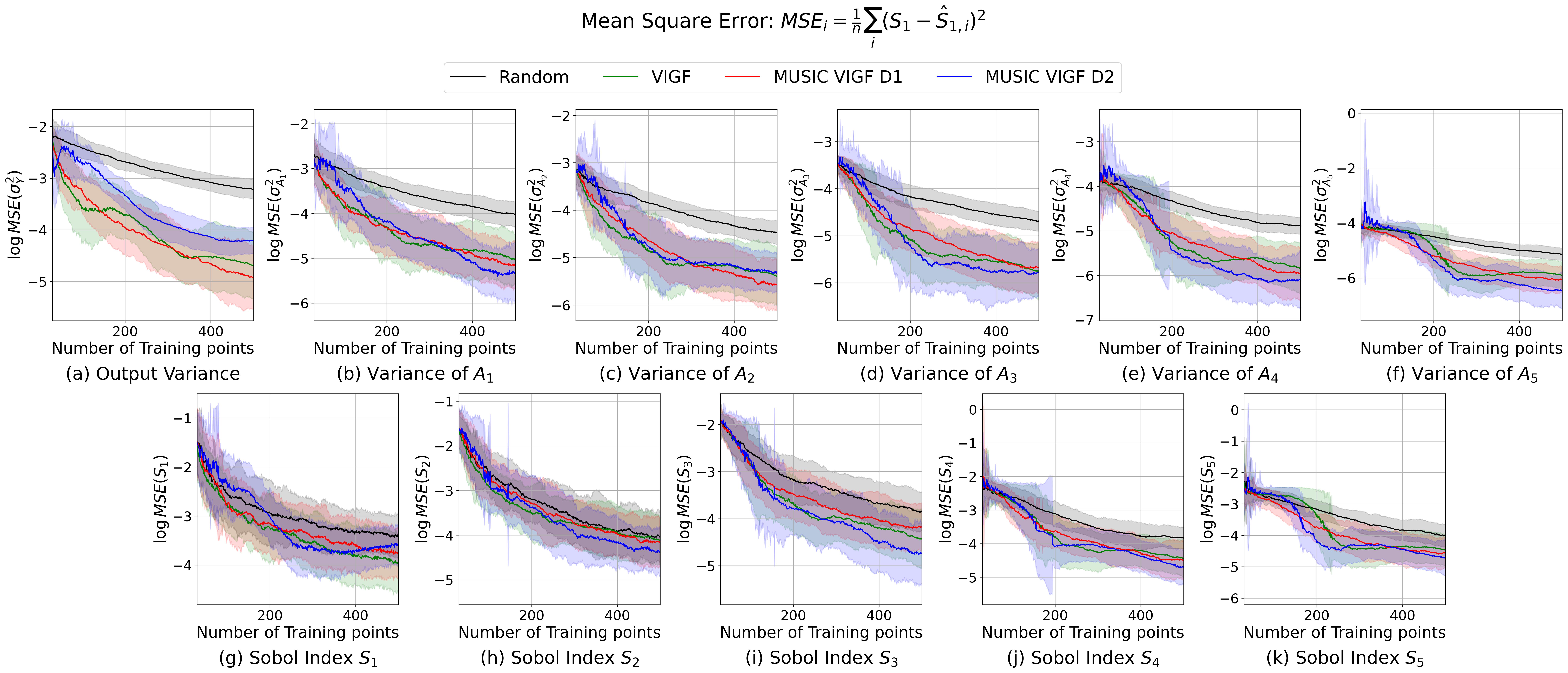}
    \caption{G-Sobol Function -- Mean square convergence with $\sigma$ confidence intervals from 100 repeated trials of (a) output variance $\sigma_Y^2$, (b--f) main effect variances $\sigma_{A_i}^2$ $\forall$ $i \in \{1, 2, 3, 4, 5\}$, and (g--k) Sobol indices $S_i$ for MUSIC and VIGF active learning schemes compared with random sampling.}
    \label{g_si_vigf}
\end{figure}

Again we compare the three sampling strategies, where each sampling strategy initiates by training a GP on 30 samples obtained from Latin Hypercube Sampling and adaptively selects new samples up to 500 samples. The Sobol indices are computed using 25000 candidate points. Figure \ref{g_si_vigf} shows the convergence for VIGF and MUSIC (based on VIGF) compared to random sampling. Here, the `MUSIC+VIGF' strategy and the VIGF show superior performance to random sampling in the total variance (Figure \ref{g_si_vigf}(a)), main effects (Figure \ref{g_si_vigf}(b)--(f)), and Sobol indices ((Figure \ref{g_si_vigf}(g)--(k)). Results from EIGF learning are not shown due to poor convergence in comparison to VIGF.

\subsection{Gaussian Function}
\begin{figure}[!ht]
    \centering
    \includegraphics[width=0.7\textwidth]{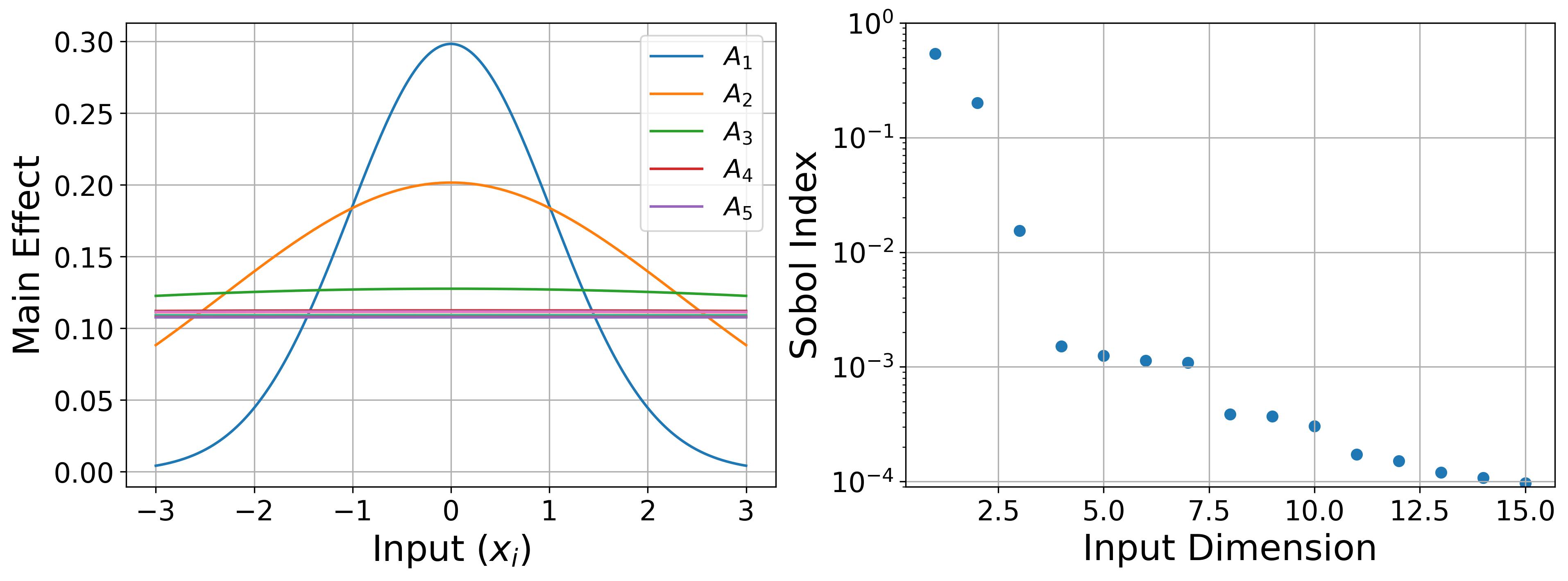}
    \caption{Gaussian function: (a) First-order main effect functions and (b) True Sobol Indices.}
    \label{fig: gaussian_function}
\end{figure}

To conclude, we compare performance for a high input dimensional problem. 
A synthetic $15$d function is considered here, such that 75.5\% of the contribution is from the first-order effects of three inputs. The function is defined as a product of 15 independent square exponential functions, similar to a Gaussian kernel given by
\begin{align*}
    y(\bm{x}) = \prod_{i=1}^{d} \exp{\Bigg( -\frac{x_i^2}{a_i}\Bigg)}
\end{align*}
where, $x_i \sim U(-3, 3)$ $\forall$ $i \in \{1, 2, \hdots, d\}$ and
$$\bm{a} = [1.45,  3.3, 15, 50, 55, 58, 59, 100, 102, 112.5, 150, 160, 180, 190, 200]$$
Figure \ref{fig: gaussian_function}(a) shows the main effect contribution and Figure \ref{fig: gaussian_function}(b) shows the true Sobol indices (in log scale).

\begin{figure}[!ht]
    \centering
    \includegraphics[width=\textwidth]{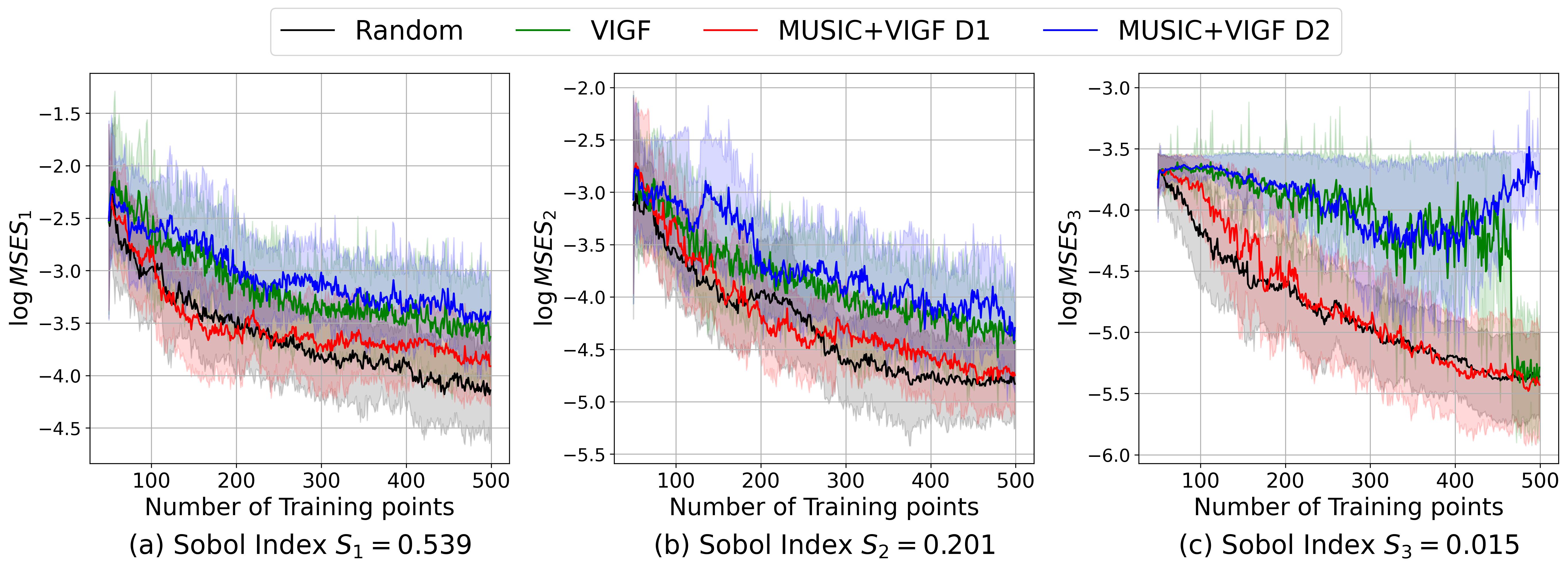}
    \caption{Gaussian Function -- Mean square convergence with $\sigma$ confidence intervals from 50 repeated trials for Sobol indices (a) $S_1$, (b) $S_2$, and (c) $S_3$ for MUSIC and VIGF active learning schemes compared with random sampling.}
    \label{fig: mg15_vigf_si}
\end{figure}

Figure~\ref{fig: mg15_vigf_si} shows convergence of the Sobol indices for the first three dimensions using the VIGF and MUSIC with VIGF learning functions compared to random sampling for 500 samples. We notice that the VIGF and `MUSIC+VIGF D2' perform quite poorly for this example. In particular, the `MUSIC+VIGF D2' exhibits poor convergence because the Euclidean distance is not a good measure of distance in 15 dimensions \cite{aggarwal2001surprising}. When this is replaced with the component-wise distance measure, convergence is much better and is comparable with random sampling. Convergence for dimensions 4--15 are not shown here because they immediately produce very low errors in all cases given that the corresponding Sobol indices are near zero. Importantly, we observe that no learning function is capable of producing superior convergence to random sampling in high-dimensions. 

\section{Application to Boundary Layer Wind Tunnel Experimental Design}
\label{sec:blwt}

\begin{figure}[!ht]
    \centering
    \includegraphics[width=\textwidth]{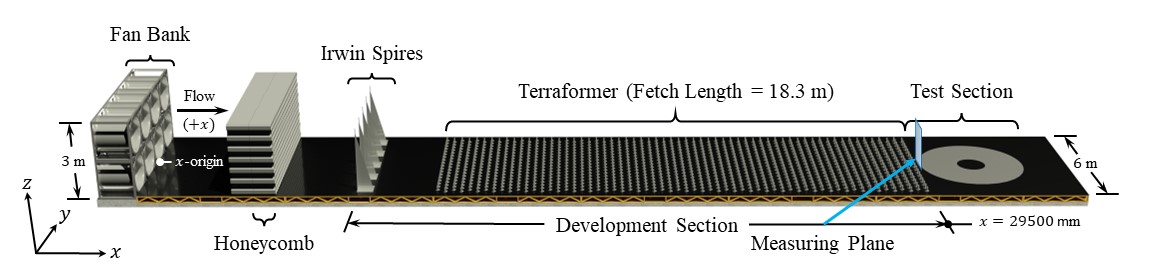}
    \caption{Schematic of the University of Florida Boundary Layer Wind Tunnel (BLWT). The Terraformer roughness element grid is capable of rapidly and automatically reconfiguring its 1116 elements to produce a wide-range of wind flow characteristics.}
    \label{terraformer}
\end{figure}

The original motivation for developing this active learning scheme for Sobol index estimation came from an experimental investigation into the influence of roughness terrain parameters in large-scale Boundary Layer Wind Tunnel (BLWT) experiments. In this study, experiments were performed using the University of Florida BLWT (UF-BLWT illustrated in Figure~\ref{terraformer}), which features a novel automated roughness element grid called the Terraformer. This Terraformer is capable of rapidly and automatically changing the roughness terrain at the push of a button to mimic a wide-range of wind flow conditions \cite{catarelli2020automation}. 

In this work, we parameterize the Terraformer (which has 1116 independent elements capable of changing both height and profile orientation) by a random field in the along-wind direction using a truncated Karhunen-Loeve (KL) expansion such that the element at a location $x$ along the Terraformer is given by
\begin{align*}
    h(x, \boldsymbol{\theta}) = \mu_H + \sum_{i=1}^{d} \sqrt{\lambda^{(i)}} {\theta}^{(i)} f^{(i)}(x)
\end{align*}
where $\mu_H=80$ mm is the mean element height, $\theta^{(i)}$ are standard normal random variables, and $\lambda^{(i)}$ and $f^{(i)}(x)$ are the eigenvalues and eigenfunctions of the covariance function given by
\begin{equation}
    C(x_1,x_2)=\sigma^2 \exp(-a|x_2 - x_1|)\cos(\omega|x_2 - x_1|)
\end{equation}
Here $\sigma^2 = 100$ mm is the variance of the field, $\omega$ is the wave number, and $a$ and $\omega$ are selected such that the length scale of the covariance function is given by $L = a/(a^22 + \omega^2) = 3000$ mm, or approximately 1/6 of the Terraformer length. The KL expansion is truncated to $d=10$ terms.

The significant expense in this project is due to experimental time, where the execution time for a single experiment is approximately 20 minutes. Even with the high degree of automation afforded by the UF-BLWT, we cannot afford to explore the influence of all $d=10$ random field parameters on the resulting wind flow. We therefore aim to adaptively perform global sensitivity analysis to identify which random field parameters have the most influence on the wind characteristic of interest and reduce the dimension of the random field to facilitate a feasible set of experiments. Our specific characteristic of interest is the turbulence intensity profile of the resulting wind flow defined at a height above the floor ($z$) by
\begin{equation}
    I_u(z) = \dfrac{\sigma_u(z)}{\mu_u(z)}
\end{equation}
where $\mu_u(z)$ and $\sigma_u(z)$ are the mean and standard deviation of the longitudinal component of the wind velocity $u(z)$ -- which is a random process. More specifically, we aim to quantify the sensitivity of the following distance measure to the 10 KL random variables:
\begin{equation}
    d(\boldsymbol{\theta}) = ||\bm{I}_u(\boldsymbol{\theta}) - \bm{I}_u^* ||_2
\end{equation}
where $\bm{I}_u(\boldsymbol{\theta}) $ is the vector of discretized turbulence intensities observed at points separated by 20 mm in the range $z=[180, 500]$ mm for a Terraformer configuration with parameters $\boldsymbol{\theta}$ and $\bm{I}_u^*$ is the same vector defined for a reference roughness grid with all element uniformly set to $h=80$ mm. In other words, we aim to quantify the sensitivity of the deviation of the turbulence intensity profile from the reference profile to each of the KL random variables. 

For this investigation, we applied an earlier component-wise application of the proposed EIGF-based MUSIC learning function not demonstrated in detail above. This initial version of MUSIC focuses on only one dimension at a time and chooses a new sample to minimize the uncertainty in the estimate of the particular Sobol index associated with the largest MUSIC component. More specifically, using the expected main effect improvement defined in Eq. \eqref{eqn:exp_sobol_improvement}, this version selects the dimension with maximum improvement in the individual main effect GPs as 
\begin{align}
   x_i^* = \argmax_{x_i} {E[I_{A_i}(x_i)]}
\end{align}
where $x_i^*$ is the $i^{\text{th}}$ element of the new sample $\bm{x^*}$ and the other elements are selected randomly over the input marginal distribution. Although this version of the MUSIC function was ultimately improved and the methods discussed in detail above were explored in more detail, it was successfully used to adaptively estimate sensitivities for the BLWT experimental investigation. 

\begin{figure}[h!]
    \centering
    \includegraphics[width=0.65\textwidth]{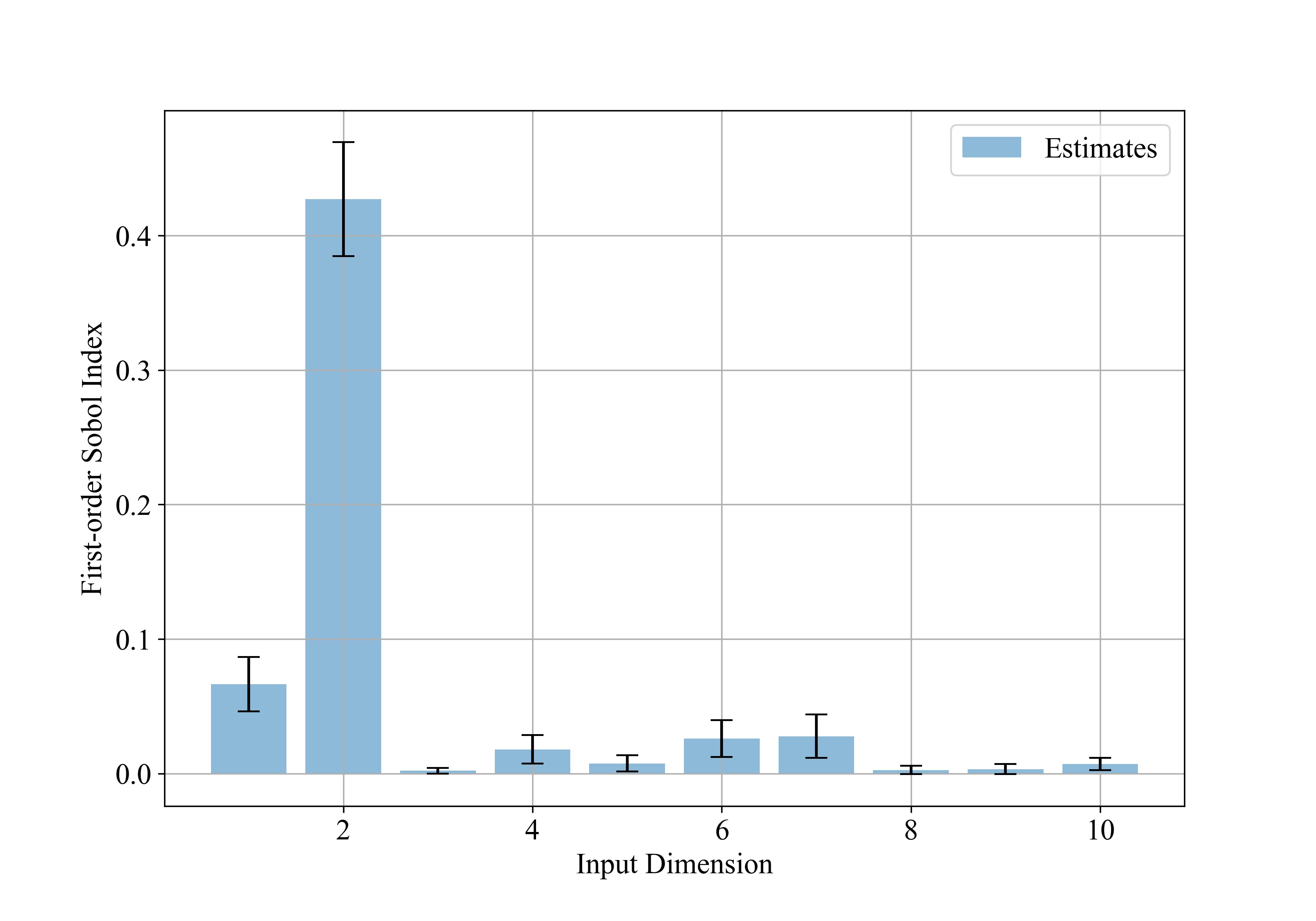}
    \caption{Final estimates of the first-order Sobol Indices.}
    \label{blwt_si}
\end{figure}

Initially, 30 experiments were performed with parameters $\boldsymbol{\theta}$ selected by Latin HyperCube sampling. Then, 270 more experiments were conducted using the MUSIC learning strategy described above. \Cref{blwt_si} shows the final estimates (after 300 experiments) of the first-order Sobol indices, along with the standard error in the estimates. The experiments show that the distance metric between two roughness configurations is dominated by the second eigen parameter. Furthermore, except for input dimensions 1 and 2, no other input parameter has a considerable first-order effect on the distance. Using the approach described in Appendix B, we were further able to make an initial assessment of the interaction sensitivities and determine that significant interactions between inputs 1, 2, and 3 play an important role in the resulting turbulence intensity profiles. Using these insights, we were able to reduce the dimension of the KL random field to $d=3$ and conduct a more rigorous study, further details of which are provided in \cite{shields2023}.

\section{Conclusions}
This study investigates the application of active learning for global sensitivity analysis. We specifically propose a novel new learning function, termed the MUSIC (minimize uncertainty in Sobol index convergence) function, that leverages the main effect variances in Gaussian Process-based Sensitivity Indices, i.e. the numerator in Sobol index definition. We compare this strategy with a more traditional active learning strategy aimed at accurately capturing output variance $\sigma^2_Y$, i.e. the denominator of the Sobol index, using the existing EIGF and VIGF learning functions. Both strategies are further compared with simple sequential random sampling. We provide a careful analysis of the convergence for four analytical test functions and study the convergence behavior. An important insight that arises is that convergence in Sobol indices are difficult for active learning because they are defined as a ratio and the learning strategies must focus on either the numerator or the denominator and achieving rapid convergence in either, or both, is not sufficient to ensure convergence in the ratio. 

Summarizing, we can make the following observations:
\begin{itemize}
    \item In general, the VIGF function outperforms the EIGF, as does the MUSIC with VIGF when compared to the MUSIC with EIGF. 
    \item The MUSIC learning functions are very successful at rapidly identifying small Sobol indices. 
    \item For low-dimensional problems, the MUSIC with VIGF and distance pre-factor based on a Euclidean distance generally outperforms the other methods in terms of Sobol index convergence.
    \item The Euclidean distance-based pre-factor ($\bm{D}_2$) is effective for low-dimensional problems but does not work well in high-dimensions.
    \item At best, the MUSIC learning strategy is comparable in performance to random sampling for high-dimensional problems.
\end{itemize}

Finally, we provide a motivating application where active learning is applied to learn global sensitivity indices for a set of large-scale boundary layer wind tunnel experiments. This demonstrates a real, practical use case and illustrates how it helped to significantly reduce experimental effort.

\subsection*{Acknowledgements}
This material is based upon work supported by the National Science Foundation under Grant Nos. 1930389 and 1930625. Dr.\ Dimitrios Tsapetis has been supported by Defense Threat Reduction Agency, Award HDTRA1202000.

\bibliographystyle{unsrtnat}
\bibliography{references}

\appendix

\renewcommand{\theequation}{A.\arabic{equation}}
\setcounter{equation}{0}

\section{Appendix 1}
\label{app:1}

In this appendix, we show how to compute the mean and covariance functions for the main effect GPs under specific conditions. We then demonstrate that under conditions of widely used correlation models, these functions can be evaluated analytically without the need for numerical integrations.

\subsection{Mean function of the main effect GPs}
\label{sec:main_effect_mean}

We begin by restating Eq.\ \eqref{eq:marginal_mean} as:
\begin{align*}
    E[A(X_{i})] = \int_{\bm{x}_{\sim i}} \hat{y}({\mathbf{\bm{x}}})\prod_{j\neq i} p_{X_{j}}(x_j) dx_j 
\end{align*}
Next, we simply plug in the expression for $\hat{y}$ from Eq.\ \eqref{eqn:predictor}, which yields
\begin{align}
    E[A(X_{i})] &= \int_{\bm{x}_{\sim i}} (\bm{f}(\bm{x})^T \boldsymbol{\beta} + \bm{r}(\bm{x})^T \bm{R}^{-1} (\bm{Y} - \bm{F}\boldsymbol{\beta}))\prod_{j\neq i} p_{X_{j}}(x_j) dx_j \\
     &= \bigg[\int_{\bm{x}_{\sim i}} \bm{f}(\bm{x})^T \prod_{j\neq i} p_{X_{j}}(x_j) dx_j \bigg]\boldsymbol{\beta} + \bigg[\int_{\bm{x}_{\sim i}} \bm{r}(\bm{x})^T \prod_{j\neq i} p_{X_{j}}(x_j) dx_j \bigg] \bm{R}^{-1} (\bm{Y} - \bm{F}\boldsymbol{\beta})
     \label{eqn:mean_expanded}
\end{align}
Integrating the first term of Equation \eqref{eqn:mean_expanded} for a linear basis $\bm{f}(\bm{x}) = \begin{bmatrix} 1 & x_1 & x_2 & \hdots & x_d\end{bmatrix}^T$.
\begin{align*}
    \int_{\bm{x}_{\sim i}} \bm{f}(\bm{x})^T \prod_{j\neq i} p_{X_{j}}(x_j) dx_j & = \int_{\bm{x}_{\sim i}} \begin{bmatrix}
1 & x_1 & x_2 & \hdots & x_d
\end{bmatrix}\prod_{j\neq i} p_{X_{j}}(x_j) dx_j \\  
& = \begin{bmatrix}
1 & E[X_1] & E[X_2] & \hdots & x_i & \hdots & E[X_d]
\end{bmatrix}
\end{align*} 
Integrating the second term of Equation \eqref{eqn:mean_expanded} yields:
\begin{align*}
    \int_{\bm{x}_{\sim i}} \bm{r}(\bm{x})^T \prod_{j\neq i} p_{X_{j}}(x_j) dx_j =& \int_{\bm{x}_{\sim i}} \begin{bmatrix} \mathcal{R}({\bm{x}}, \bm{x}^{(1)}) \\ \mathcal{R}({\bm{x}}, \bm{x}^{(2)}) \\ \vdots \\ \mathcal{R}({\bm{x}}, \bm{x}^{(n)}) \end{bmatrix}^T \prod_{j\neq i} p_{X_{j}}(x_j) dx_j \\ 
     =& \int_{\bm{x}_{\sim i}} \begin{bmatrix} \prod_{k}r_k(x_{k}, x_k^{(1)}) \\ \prod_{k} r_k(x_{k}, x_k^{(2)}) \\ \vdots \\ \prod_{k} r_k(x_{k}, x_k^{(n)}) \end{bmatrix}^T \prod_{j\neq i} p_{X_{j}}(x_j) dx_j\\
    =&  \begin{bmatrix} r_i(x_{i}, x_i^{(1)}) \prod_{j\neq i} \int_{x_j} r_j(x_{j}, x_j^{(1)})p_{X_{j}}(x_j) dx_j \\ r_i(x_{i}, x_i^{(2)}) \prod_{j\neq i} \int_{x_j} r_j(x_{j}, x_j^{(2)})p_{X_{j}}(x_j) dx_j \\ \vdots \\ r_i(x_{i}, x_i^{(n)}) \prod_{j\neq i} \int_{x_j} r_j(x_{j}, x_j^{(n)})p_{X_{j}}(x_j) dx_j \end{bmatrix}^T 
\end{align*}
where we recall that $\bm{x}^{(l)}, \sim l=1,\dots,n$ are the training samples and $r_k(x_k,x_k^{(l)})$ is the $k^{th}$ univariate correlation function expressed in Eq.\ \eqref{eqn:correlation}.
Finally, Eq.\ \eqref{eqn:mean_expanded} can be expressed as:
\begin{align}
\label{simp_mean_vec}
E[A(X_{i})] = \begin{bmatrix}
1 \\ E[X_1] \\ E[X_2] \\ \vdots\\ x_{i} \\ \vdots \\  E[X_d]]
\end{bmatrix}^T \boldsymbol{\beta} + \begin{bmatrix} r_i(x_{i}, x_i^{(1)}) \prod_{j\neq i} \int_{x_j} r_j(x_{j}, x_j^{(1)})p_{X_{j}}(x_j) dx_j \\ r_i(x_{i}, x_i^{(2)}) \prod_{j\neq i} \int_{x_j} r_j(x_{j}, x_j^{(2)})p_{X_{j}}(x_j) dx_j \\ \vdots \\ r_i(x_{i}, x_i^{(n)}) \prod_{j\neq i} \int_{x_j} r_j(x_{j}, x_j^{(n)})p_{X_{j}}(x_j) dx_j \end{bmatrix}^T  \bm{R}^{-1} (\bm{Y} - \bm{F}\boldsymbol{\beta})
\end{align}
Finally, in Section \ref{subsec:correlation_integration}, we show that the product of 1D integrals can be computed analytically for the Gaussian correlation model, which results in the following closed form for the mean of the main effect GP for uniformly distributed $X$:
\begin{align}
\label{simp_mean_vec1}
E[A&(X_{i})] = \begin{bmatrix}
1 & E[X_1] & E[X_2] & \hdots & x_{i} & \hdots &  E[X_d]]
\end{bmatrix} \boldsymbol{\beta} \nonumber \\ 
& + \begin{bmatrix} r_i(x_{i}, x_i^{(1)}) \prod_{j\neq i} \frac{1}{b_j-a_j} \sqrt{\frac{\pi}{\theta_j}} [ \Phi(\sqrt{2 \theta_j} (b_j-{x_j^{(1)}})) - \Phi(\sqrt{2 \theta_j} (a_j-{x_j^{(1)}}))] \\ r_i(x_{i}, x_i^{(2)}) \prod_{j\neq i} \frac{1}{b_j-a_j} \sqrt{\frac{\pi}{\theta_j}} [ \Phi(\sqrt{2 \theta_j} (b_j-{x_j^{(2)}})) - \Phi(\sqrt{2 \theta_j} (a_j-{x_j^{(2)}}))] \\ \vdots \\ r_i(x_{i}, x_i^{(n)}) \prod_{j\neq i} \frac{1}{b_j-a_j} \sqrt{\frac{\pi}{\theta_j}} [ \Phi(\sqrt{2 \theta_j} (b_j-{x_j^{(n)}})) - \Phi(\sqrt{2 \theta_j} (a_j-{x_j^{(n)}}))] \end{bmatrix}^T  \bm{R}^{-1} (\bm{Y} - \bm{F}\boldsymbol{\beta})
\end{align}

Note that in cases where the GP is trained with noisy data the above equation can be easily modified to incorporate the noise. We simply substitute the correlation matrix, $\bm{R}$, in the above equation by $\bm{R_n}$ defined as:
\begin{align*}
    \bm{R}_n = ((1-\tau)\bm{R}+\tau \bm{I})
\end{align*}
where ${\tau=\sigma_{\epsilon}^2}/{\sigma_{z}^2}$ and $\sigma_{\epsilon}^2$ is the variance due to the noise in the data and $\sigma_{z}^2$ is the total variance of the GP. 

\subsection{Covariance function of the main effect GPs}
\label{sec:main_effect_cov}

We again begin by restating the expression for the covariance function of the main effect GPs from Eq. \eqref{eq:marginal_cov}:
\begin{align*}
    \text{Cov}(A(X_{1i}), A(X_{2i})) & = \int_{\bm{x}_{1\sim i}} \int_{\bm{x}_{2\sim i}} \text{Cov}(\mathcal{Y}(\bm{x}_1), \mathcal{Y}(\bm{x}_2)) \prod_{j\neq i} p_{X_j}(x_{1j}) d{x_{1j}} \prod_{k\neq i} p_{X_k}(x_{2k}) d{x_{2k}}
\end{align*}
The covariance of the GP $\mathcal{Y}(\bm{x})$ can be expressed as:
\begin{align*}
    \text{Cov}(\mathcal{Y}(\bm{x}_1), \mathcal{Y}(\bm{x}_2)) = \sigma^2_z (r(\bm{x}_{1}, \bm{x}_{2})-\bm{r}(\bm{x}_{1})^T \bm{R}^{-1} \bm{r}(\bm{x}_{2}) + \bm{t}(\bm{x}_{1})^T( \bm{F}^T \bm{R}^{-1} \bm{F})^{-1} \bm{t}(\bm{x}_{2}) ),
\end{align*}
which yields the following covariance for the main effect GP:
\begin{multline}
    \text{Cov}(A(X_{1i}), A(X_{2i}))  = \int_{\bm{x}_{1\sim i}} \int_{\bm{x}_{2\sim i}} \bigg[ \sigma^2_z (r(\bm{x}_{1}, \bm{x}_{2})-\bm{r}(\bm{x}_{1})^T \bm{R}^{-1} \bm{r}(\bm{x}_{2}) + \bm{t}(\bm{x}_{1})^T( \bm{F}^T \bm{R}^{-1} \bm{F})^{-1} \bm{t}(\bm{x}_{2}) ) \bigg] \\ \prod_{j\neq i} p_{X_j}(x_{1j}) d{x_{1j}} \prod_{k\neq i} p_{X_k}(x_{2k}) d{x_{2k}}
    \label{eqn:main_effect_cov}
\end{multline}
Expanding this, we can express the first term of Eq.\ \eqref{eqn:main_effect_cov} as:
\begin{align*}
\label{}
    \sigma^2_z \int_{\bm{x}_{1\sim i}} \int_{\bm{x}_{2\sim i}} r(\bm{x}_{1}, \bm{x}_{2}) \prod_{j\neq i} p_{X_j}(x_{1j}) d{x_{1j}} \prod_{k\neq i} p_{X_k}(x_{2k}) d{x_{2k}} = & \\
    \sigma^2_z \int_{\bm{x}_{1\sim i}} \int_{\bm{x}_{2\sim i}} \prod_{l=1}^d r_l(x_{1l}, x_{2l}) \prod_{j\neq i}  & p_{X_j}(x_{1j}) d{x_{1j}} \prod_{k\neq i} p_{X_k}(x_{2k}) d{x_{2k}} & \\
    = \sigma^2_z r_i(x_{1i}, x_{2i}) \prod_{l \neq i}^d \int_{\bm{x}_{1\sim i}}  \int_{\bm{x}_{2\sim i}}  r_l(x_{1l} &, x_{2l})  p_{X_l}(x_{1l}) d{x_{1l}}   p_{X_l}(x_{2l}) d{x_{2l}}
\end{align*}
In Section \ref{subsec:correlation_integration}, we derive the closed form for this integral given a uniform distribution and Gaussian correlation model, which yields:
\begin{multline*}
    \sigma^2_z \int_{\bm{x}_{1\sim i}} \int_{\bm{x}_{2\sim i}} r(\bm{x}_{1}, \bm{x}_{2}) \prod_{j\neq i} p_{X_j}(x_{1j}) d{x_{1j}} \prod_{k\neq i} p_{X_k}(x_{2k}) d{x_{2k}} =  \\
    \sigma^2_z r_i(x_{1i}, x_{2i}) \prod_{l \neq i}^d  \left(\frac{1}{b_l-a_l}\right)^2 \sqrt{\frac{\pi}{\theta_l}} \times \\ \left[ a_l + b_l - 2 [a_l \Phi(\sqrt{2 \theta_l}(b_l-a_l)) + b_l \Phi(\sqrt{2 \theta_l}(a_l - b_l))] - \frac{1}{\sqrt{\pi \theta_l}} [1 - \exp \{ -\theta_l(b_l-a_l)^2\}] \right] 
\end{multline*}
Next, the second term of the Eq.\ \eqref{eqn:main_effect_cov} is:
\begin{align*}
   -\sigma^2_z \int_{\bm{x}_{1\sim i}} \int_{\bm{x}_{2\sim i}}  \bm{r}(\bm{x}_{1})^T \bm{R}^{-1} \bm{r}(\bm{x}_{2}) \prod_{j\neq i} p_{X_j}(x_{1j}) d{x_{1j}} \prod_{k\neq i} p_{X_k}(x_{2k}) d{x_{2k}} = & \\
   -\sigma^2_z \bigg[ \int_{\bm{x}_{1\sim i}} \bm{r}(\bm{x}_{1})^T \prod_{j\neq i} p_{X_j}(x_{1j}) d{x_{1j}}  \bigg] \bm{R}^{-1} \bigg[ \int_{\bm{x}_{2\sim i}} \bm{r}(\bm{x}_{2}) &  \prod_{k\neq i} p_{X_k}(x_{2k}) d{x_{2k}} \bigg]
\end{align*}
where the integrals in brackets can be solved in closed form as derived in Section \ref{sec:main_effect_mean}.
Finally, the third term of Eq.\ \eqref{eqn:main_effect_cov} is:
\begin{align*}
   -\sigma^2_z \int_{\bm{x}_{1\sim i}} \int_{\bm{x}_{2\sim i}}  \bm{t}(\bm{x}_{1})^T( \bm{F}^T \bm{R}^{-1} \bm{F})^{-1} \bm{t}(\bm{x}_{2}) \prod_{j\neq i} p_{X_j}(x_{1j}) d{x_{1j}} \prod_{j\neq i} p_{X_j}(x_{2j}) d{x_{2j}} = & \\
   -\sigma^2_z \bigg[ \int_{\bm{x}_{1\sim i}} \bm{t}(\bm{x}_{1})^T \prod_{j\neq i} p_{X_j}(x_{1j}) d{x_{1j}}  \bigg] (\bm{F}^T \bm{R}^{-1} \bm{F})^{-1} \bigg[ \int_{\bm{x}_{2\sim i}} \bm{t}(\bm{x}_{2})&  \prod_{j\neq i} p_{X_j}(x_{2j}) d{x_{2j}} \bigg]
\end{align*}
where, $\hat{\sigma}^2_z$ is the variance of the GP, ${\mathcal{Y}}$, and $\bm{t}(\bm{x})=\bm{F}^T\bm{R}^{-1}\bm{r}(\bm{x})-\bm{f}(\bm{x})$ as defined in  Eq.\ \eqref{eqn:krig_variable_t}. Finally, the integral of $\bm{t}(\bm{x})$ over the joint distribution in brackets can be easily obtained from the equations in Section \ref{sec:main_effect_mean}, as follows
\begin{align*}
    \int_{\bm{x}_{1\sim i}} \bm{t}(\bm{x})^T  \prod_{j\neq i} p_{X_j}(x_{j}) d{x_{j}} &= \int_{\bm{x}_{1\sim i}} \Big[ \bm{F}^T\bm{R}^{-1}\bm{r}(\bm{x})-\bm{f}(\bm{x}) \Big] \prod_{j\neq i} p_{X_j}(x_{j}) d{x_{j}} \\
    &= \bm{F}^T\bm{R}^{-1} \int_{\bm{x}_{1\sim i}} \bm{r}(\bm{x}) \prod_{j\neq i} p_{X_j}(x_{j}) d{x_{j}} - \int_{\bm{x}_{1\sim i}} \bm{f}(\bm{x}) \prod_{j\neq i} p_{X_j}(x_{j}) d{x_{j}} \\
    &=\bm{F}^T\bm{R}^{-1}\begin{bmatrix} r_i(x_{i}, x_i^{(1)}) \prod_{j\neq i} \int_{x_j} r_j(x_{j}, x_j^{(1)})p_{X_{j}}(x_j) dx_j \\ r_i(x_{i}, x_i^{(2)}) \prod_{j\neq i} \int_{x_j} r_j(x_{j}, x_j^{(2)})p_{X_{j}}(x_j) dx_j \\ \vdots \\ r_i(x_{i}, x_i^{(n)}) \prod_{j\neq i} \int_{x_j} r_j(x_{j}, x_j^{(n)})p_{X_{j}}(x_j) dx_j \end{bmatrix}  - \begin{bmatrix}
1 \\ E[X_1] \\ E[X_2] \\ \vdots\\ x_{i} \\ \vdots \\  E[X_d]]
\end{bmatrix}
\end{align*}
The above expression can be further simplified for a given uniform distribution and Gaussian correlation model as:
\begin{align*}
    \int_{\bm{x}_{1\sim i}} \bm{t}(\bm{x})^T & \prod_{j\neq i} p_{X_j}(x_{j}) d{x_{j}}  \\
    & = \bm{F}^T\bm{R}^{-1} \begin{bmatrix} r_i(x_{i}, x_i^{(1)}) \prod_{j\neq i} \frac{1}{b_j-a_j} \sqrt{\frac{\pi}{\theta_j}} [ \Phi(\sqrt{2 \theta_j} (b_j-{x_j^{(1)}})) - \Phi(\sqrt{2 \theta_j} (a_j-{x_j^{(1)}}))] \\ r_i(x_{i}, x_i^{(2)}) \prod_{j\neq i} \frac{1}{b_j-a_j} \sqrt{\frac{\pi}{\theta_j}} [ \Phi(\sqrt{2 \theta_j} (b_j-{x_j^{(2)}})) - \Phi(\sqrt{2 \theta_j} (a_j-{x_j^{(2)}}))] \\ \vdots \\ r_i(x_{i}, x_i^{(n)}) \prod_{j\neq i} \frac{1}{b_j-a_j} \sqrt{\frac{\pi}{\theta_j}} [ \Phi(\sqrt{2 \theta_j} (b_j-{x_j^{(n)}})) - \Phi(\sqrt{2 \theta_j} (a_j-{x_j^{(n)}}))] \end{bmatrix} - \begin{bmatrix}
1 \\ E[X_1] \\ E[X_2] \\ \vdots\\ x_{i} \\ \vdots \\  E[X_d]]
\end{bmatrix}
\end{align*}

Thus, the main effect GP, $A(X_{i})$, can be completely defined by the mean function, $E[A(X_i)]$, and covariance function, $\text{Cov}(A(X_{1i})$, $A(X_{2i}))$, derived in Sections \ref{sec:main_effect_mean} and \ref{sec:main_effect_cov}. 

Again in case of noisy training data, we simply substitute the correlation matrix, $\bm{R}$, in the above equation by $\bm{R_n}$ as defined below:
\begin{align*}
    \bm{R}_n = ((1-\tau)\bm{R}+ \tau \bm{I})
\end{align*}
where, ${\tau=\sigma_{\epsilon}^2}/{\sigma_{z}^2}$ and $\sigma_{\epsilon}^2$ is the variance due to the noise in the data and $\sigma_{z}^2$ is the total variance of the GP.

\subsection{Closed-form integration of the Gaussian correlation model}
\label{subsec:correlation_integration}

Let $x_1$ and $x_2$ be univariate sample points drawn from a uniform distribution over the range $(a,b)$, $X_1, X_2\sim U(a,b)$, having Gaussian correlation (from Eq.\ \eqref{eqn:correlation}) given by:
\begin{align*}
    r(x_1, x_2|\theta) = \exp \{-\theta(x_1-x_2)^2\} 
\end{align*}
Note that, without loss of generality, the arbitrary sample points can be transformed to a uniform distribution through a suitable isoprobabilistic transformation (e.g.\ Nataf \cite{lebrun2009innovating}). As discussed in the previous appendices, we need to integrate this correlation model over the pdf $p_{X_1}(x_1)$ as follows: 
\begin{align*}
    \int_{-\infty}^{\infty} r(\theta, x_1, x_2) p_{{X_1}}({x_1}) dx_1 = & \int_{a}^{b} \exp \{-\theta({x_1}-{x_2})^2\}  \frac{1}{b-a} d{x_1} \\
    = & \frac{1}{b-a} \int_{a}^{b} \exp \{-\theta({x_1}-{x_2})^2\}  d{x_1} \\
    = & \frac{1}{b-a} \sqrt{2 \pi (1/2\theta)} \int_{a}^{b} \frac{1}{\sqrt{2 \pi (1/2\theta)}} \exp \bigg\{-\frac{1}{2} \frac{({x_1}-{x_2})^2}{(1/2\theta)}\bigg\}  d{x_1} \\
    = & \frac{1}{b-a} \sqrt{\frac{\pi}{\theta}} \bigg[ \Phi\bigg(\sqrt{2 \theta} (b-{x_2})\bigg) - \Phi\bigg(\sqrt{2 \theta} (a-{x_2})\bigg) \bigg]
\end{align*}
where, $\Phi(\cdot)$ is the standard normal cumulative distribution function. Further, we need the integrate the above result with respect to ${x_2}$ as
\begin{equation}
\begin{aligned}
    \int_{-\infty}^{\infty} \int_{-\infty}^{\infty}  r(\theta, {x_1}, {x_2})  & p_{{X_1}}({x_1}) p_{{X_2}}({x_2})  d{x_1} d{x_2} \\
    = & \int_{-\infty}^{\infty} \bigg[ \frac{1}{b-a} \sqrt{\frac{\pi}{\theta}} \bigg[ \Phi\bigg(\sqrt{2 \theta} (b-{x_2})\bigg) - \Phi\bigg(\sqrt{2 \theta} (a-{x_2})\bigg) \bigg] \bigg] p_{{X_2}}({x_2}) d{x_2} \\
    = & \bigg(\frac{1}{b-a}\bigg)^2 \sqrt{\frac{\pi}{\theta}} \int_{a}^{b} \bigg[ \Phi\bigg(\sqrt{2 \theta} (b-{x_2})\bigg) - \Phi\bigg(\sqrt{2 \theta} (a-{x_2})\bigg) \bigg]  d{x_2} \\
\end{aligned}
\label{eqn:double_int}
\end{equation}
We see that the above integral has two identical components that differ only by the constant $a$ or $b$. Let us introduce a dummy variable, $z$, which can take values $a$ or $b$. Integration by parts yields 
\begin{align*}
    \int_{a}^{b} (1) \times \Phi & \bigg( \sqrt{2\theta} ( z - {x_2}  )\bigg) d{x_2}
    = \Phi\bigg( \sqrt{2\theta} ( z - {s})\bigg) x_2 \bigg{|}_{a}^{b} - \int_{a}^{b} {x_2} \phi \bigg( \sqrt{2\theta} ( z - {x_2}  )\bigg) (-\sqrt{2\theta}) d{x_2}
\end{align*}
where, $\phi(.)$ is the standard normal probability density function. In this integration by parts and with the use of a change of variables as $t=2\theta(z-x_2)^2$, the second term can be expressed as:
\begin{align*}
\int_{a}^{b} {x_2} \phi \bigg( \sqrt{2\theta} & ( z - {x_2}  )\bigg) (-\sqrt{2\theta}) d{x_2} \\ 
    = & \int_{a}^{b} [z - (z - {x_2})](-\sqrt{2\theta}) \frac{1}{\sqrt{2 \pi}} \exp \bigg\{ -\frac{1}{2} \frac{(z-{x_2})^2}{(1/2 \theta)}\bigg\} d {x_2} \\
    = & - z \int_{a}^{b}\frac{1}{\sqrt{2 \pi (1/2 \theta)}} \exp \bigg\{ -\frac{1}{2} \frac{(z-{x_2})^2}{(1/2 \theta)}\bigg\} d {x_2} + \int_{a}^{b} \frac{(z - {x_2})}{(1/\sqrt{2\theta})}  \frac{1}{\sqrt{2 \pi}} \exp \bigg\{ -\frac{1}{2} \frac{(z-{x_2})^2}{(1/2 \theta)}\bigg\} d {x_2}\\
    = & -z \Phi\bigg( \sqrt{2 \theta}({x_2} - z) \bigg) \bigg|_{a}^{b}  +\int_{2\theta(z-a)^2}^{2\theta(z-b)^2} \sqrt{t} \frac{1}{\sqrt{2 \pi}} \exp\bigg\{-\frac{1}{2}t\bigg\} \frac{dt}{2 \sqrt{t} (-\sqrt{2 \theta})} \\
    = & -z \Phi\bigg( \sqrt{2 \theta}({x_2} - z) \bigg) \bigg|_{a}^{b} - \frac{1}{4\sqrt{\pi \theta}} \int_{2\theta(z-a)^2}^{2\theta(z-b)^2} \exp\bigg\{-\frac{1}{2}t\bigg\} dt \\
    = & -z \Phi\bigg( \sqrt{2 \theta}({x_2} - z) \bigg) \bigg|_{a}^{b} + \frac{1}{2\sqrt{\pi \theta}}  \exp\bigg\{-\frac{1}{2}t\bigg\} \bigg|_{2\theta(z-a)^2}^{2\theta(z-b)^2}
\end{align*}
Using this expression the integral in Eq.\ \eqref{eqn:double_int}, can be evaluate as:
\begin{align*}
    \int_{a}^{b} \Phi & \bigg( \sqrt{2\theta} ( b - {x_2}  )\bigg) d{x_2} -
    \int_{a}^{b} \Phi \bigg( \sqrt{2\theta} ( a - {x_2}  )\bigg) d{x_2}  \\ & = a + b - 2 [a \Phi(\sqrt{2 \theta}(b-a)) + b \Phi(\sqrt{2 \theta}(a - b))] - \frac{1}{\sqrt{\pi \theta}} [1 - \exp \{ -\theta(b-a)^2\}]
\end{align*}
Finally, we can compute the double integration of the correlation function in closed form as:
\begin{multline}
    \int_{-\infty}^{\infty} \int_{-\infty}^{\infty} r(\theta, {x_1}, {x_2})  p_{{X_1}}({x_1}) p_{{X_2}}({x_2})  d{x_1} d{x_2}
    = \\ \bigg(\frac{1}{b-a}\bigg)^2 \sqrt{\frac{\pi}{\theta}} \bigg[ a + b - 2 [a \Phi(\sqrt{2 \theta}(b-a)) + b \Phi(\sqrt{2 \theta}(a - b))] - \frac{1}{\sqrt{\pi \theta}} [1 - \exp \{ -\theta(b-a)^2\}] \bigg] 
\end{multline}

\section{Interaction Sensitivities}
\label{app:2}
The interaction sensitivities are computed using a similar procedure as the main effect sensitivities. These indices can be expressed using following equation 
\begin{equation}
    \tilde{S}_{\bm{p}}(\omega) = \dfrac{\text{Var}_{X_{\bm{p}}}(E_{X_{\sim \bm{p}}}[\mathcal{Y}(X,\omega)|X_{\bm{p}}])}{E[\text{Var}(\mathcal{Y}(X,\omega))]}
    \label{eqn:Sobol_RV2}
\end{equation}
where, $\bm{p}$ is a subset of indices of the input dimensions(i.e. $\bm{p} \in \{1, 2, \hdots, d\}$).

To compute the interaction Sobol indices, we start by taking the expectation of the GP $\mathcal{Y}(X,\omega)$ over all the inputs except `$X_{\bm{p}}$' and denote this as
\begin{equation}
    A(X_{\bm{p}},\omega) = E_{X_{\sim \bm{p}}}[\mathcal{Y}(X,\omega)|X_{\bm{p}}]
\end{equation}
Since, $\mathcal{Y}(X,\omega)$ is a Gaussian process and expectation is a linear operator, $A(X_{\bm{p}},\omega)$ is also a GP referred to as the interaction effect GP.

The mean and covariance function of the interaction effect GP can be determined by integrating the original GP with respect to the joint probability measure over all inputs except $X_{\bm{p}}$. Considering independent inputs, the mean function is given by \cite{Oakley}: 
\begin{align}
    \mu_{A_{\bm{p}}}(x_{\bm{p}}) = \mathbb{E}[A(X_{\bm{p}})] = \int_{\bm{x}_{\sim {\bm{p}}}} \hat{y}(\bm{x})\prod_{j\not\in {\bm{p}}} p_{X_j}(x_j) dx_j 
    \label{eq:marginal_mean2}
\end{align}
which is easily computed as a product of one-dimensional integrals. Again considering independent inputs, the covariance function of the interaction effect GP is given by \cite{Oakley}:
\begin{align}
    \text{Cov}(A(X_{1{\bm{p}}}), A(X_{2{\bm{p}}})) & = \int_{\bm{x}_{1\sim {\bm{p}}}} \int_{\bm{x}_{2\sim {\bm{p}}}} \text{Cov}(\mathcal{Y}(\bm{x}_1), \mathcal{Y}(\bm{x}_2)) \prod_{j\not\in {\bm{p}}} p_{X_j}(x_{1j}) d{x_{1j}} \prod_{j\not\in {\bm{p}}} p_{X_j}(x_{2j}) d{x_{2j}}
    \label{eq:marginal_cov2}
\end{align}
which can be expressed as a product of 2-dimensional integrals. After obtaining the mean and covariance function of the interaction effect GP, the estimates for sensitivities indices are obtained in a similar procedure as main effect sensitivities. Note that the contributions of the interaction sensitivities are not used during the learning process.
\end{document}